\documentclass[journal]{IEEEtai}

\usepackage[colorlinks,urlcolor=blue,linkcolor=blue,citecolor=blue]{hyperref}

\usepackage{color,array}

\usepackage{graphicx}
\usepackage{microtype}
\usepackage{subfigure}
\usepackage{times}  % DO NOT CHANGE THIS
\usepackage{helvet}  % DO NOT CHANGE THIS
\usepackage{courier}  % DO NOT CHANGE THIS
\usepackage{caption} % DO NOT CHANGE THIS AND DO NOT ADD ANY OPTIONS TO IT
\usepackage{booktabs}
\usepackage{multirow}
\usepackage{bm} 
\usepackage{tcolorbox}
\usepackage{placeins}
\usepackage{fancybox}
\usepackage{svg}            
\usepackage{booktabs} % for professional tables
\usepackage{algorithm}
\usepackage{algorithmic}
\usepackage{siunitx}
\usepackage{amsmath}
\usepackage{amssymb}
\usepackage{amsfonts} % 如果你不用 amssymb，也可单独加这个
\usepackage{hyperref}   % 论文一般最终都会用
\usepackage{newfloat}
\usepackage{listings}
\usepackage{xcolor}
\usepackage{listings}
\usepackage{tcolorbox}
\tcbuselibrary{listings,breakable,skins}
\usepackage{tabularx}
\usepackage{array}
\usepackage{siunitx}
\usepackage{url}

\newcolumntype{L}{>{\raggedright\arraybackslash}X}

\definecolor{pe}{RGB}{255,235,235}
\definecolor{hanblue}{rgb}{0.27, 0.42, 0.81}

\definecolor{codebg}{RGB}{255,255,255}      % white background
\definecolor{codeframe}{RGB}{200,200,200}   % light gray border
\definecolor{codecomment}{RGB}{0,128,0}     % green comments
\definecolor{codekeyword}{RGB}{140,0,140}   % purple-ish keywords
\definecolor{codestring}{RGB}{180,0,0}      % dark red strings
\definecolor{codenumber}{RGB}{0,110,140}    % teal-ish numbers
\definecolor{codeerror}{RGB}{200,0,0}       % red for error annotations

\lstdefinestyle{pychrono-light}{
  language=Python,
  basicstyle=\ttfamily\scriptsize\color{black},
  backgroundcolor=\color{codebg},
  frame=none,
  numbers=none,
  breaklines=true,
  breakatwhitespace=true,
  showstringspaces=false,
  tabsize=4,
  keepspaces=true,
  keywordstyle=\color{codekeyword}\bfseries,
  commentstyle=\color{codecomment},
  stringstyle=\color{codestring},
  columns=fullflexible,
  morekeywords={chrono,irr,veh,sens},
  escapeinside={(*@}{@*)}
}

\lstdefinestyle{prompt}{
  language={},
  basicstyle=\ttfamily\footnotesize\color{black},
  backgroundcolor=\color{white},
  numbers=none,
  frame=none,
  breaklines=true,
  breakatwhitespace=true,
  showstringspaces=false,
  keepspaces=true,
  tabsize=2,
  columns=fullflexible,
  aboveskip=0pt,
  belowskip=0pt,
  emphstyle=\bfseries\color{black},
  emph={
    Turn,Turn~1,Turn~2,Turn~3,
    Vague,Sharp,Request,Requirements,
    PyChrono,Chrono,API,documentation,
    Code,Error,Correctness,Completeness,Efficiency,
    Sensor,LiDAR,IMU,GPS,Manager,Update
  },
  escapeinside={(*@}{@*)}
}

\lstdefinestyle{diff}{
  basicstyle=\ttfamily\scriptsize\color{black},
  backgroundcolor=\color{white},
  numbers=none,
  frame=none,
  breaklines=true,
  columns=fullflexible,
  showstringspaces=false,
  escapeinside={(*@}{@*)}
}

\newtcblisting{PromptBlock}[2][]{%
  listing only,
  listing options={style=prompt},
  breakable,
  enhanced,
  colback=white,
  colframe=black!15,
  boxrule=0.4pt,
  arc=1pt,
  left=4pt,right=4pt,top=2pt,bottom=2pt,
  title={#2},
  fonttitle=\bfseries\footnotesize,
  #1
}

\newtcblisting{CodeBlock}[2][]{%
  listing only,
  listing options={style=pychrono-light},
  enhanced,
  colback=white,
  colframe=black!15,
  boxrule=0.4pt,
  arc=1pt,
  left=4pt,right=4pt,top=2pt,bottom=2pt,
  title={#2},
  fonttitle=\bfseries\footnotesize,
  #1
}

\newtcblisting{DiffBlock}[2][]{%
  listing only,
  listing options={style=diff},
  enhanced,
  colback=white,
  colframe=black!15,
  boxrule=0.4pt,
  arc=1pt,
  left=4pt,right=4pt,top=2pt,bottom=2pt,
  title={#2},
  fonttitle=\bfseries\footnotesize,
  #1
}

\newcommand{\noindentBold}[1]{\par\noindent\textbf{#1}}

\graphicspath{{images/},{../../../../../local-image-archive/AI/LLMs/}}
\begin{document}

\title{SimBench: A Framework for Evaluating and
Diagnosing LLM-Based Digital-Twin Generation
for Multi-Physics Simulation}

\author{Jingquan~Wang, Andrew~Negrut, Hongyu~Wang, Harry~Zhang, and Dan~Negrut,~\IEEEmembership{Member,~IEEE}%
\thanks{Jingquan~Wang, Harry~Zhang, and Dan~Negrut are with the Department of Mechanical Engineering, University of Wisconsin--Madison, Madison, WI, USA.}%
\thanks{Hongyu~Wang is with the Department of Electrical and Computer Engineering, University of Wisconsin--Madison, Madison, WI, USA.}%
\thanks{Andrew~Negrut is with the Department of Computer Science, Rice University, Houston, TX, USA.}%
\thanks{This work is supported in part by the National Science Foundation under grant CMMI2153855.}
\thanks{Corresponding author: Dan~Negrut} % uncomment if you want this
\thanks{E-mail: Dan~Negrut (negrut@wisc.edu).}
}

\markboth{Journal of IEEE Transactions on Artificial Intelligence, Vol. 00, No. 0, Month 2020}
{Jingquan Wang \MakeLowercase{\textit{et al.}}: SimBench: Evaluating and Diagnosing Multi-Physics Digital Twin Code Generation via Multi-Turn LLM Interactions}

\maketitle

\begin{abstract}
We introduce SimBench, a benchmark designed to evaluate the proficiency of simulator-oriented LLMs (S-LLMs) in generating digital twins (DTs) that can be used in simulators for virtual testing. Given a collection of S-LLMs, this benchmark ranks them according to their ability to produce high-quality DTs. We demonstrate this by comparing over 33 open- and closed-source S-LLMs. Using multi-turn interactions, SimBench employs an LLM-as-a-judge (J-LLM) that leverages both predefined rules and human-in-the-loop guidance to assign scores for the DTs generated by the S-LLM, thus providing a consistent and expert-inspired evaluation protocol. The J-LLM is specific to a simulator, and herein the proposed benchmarking approach is demonstrated in conjunction with the open-sourceChrono multi-physics simulator. Chrono provided the backdrop used to assess an S-LLM in relation to the latter's ability to create digital twins for multibody dynamics, finite element analysis, vehicle dynamics, robotic dynamics, and sensor simulations. The proposed benchmarking principle is broadly applicable and enables the assessment of an S-LLM's ability to generate digital twins for other simulation packages, e.g., ANSYS, ABAQUS, OpenFOAM, StarCCM+, IsaacSim, and pyBullet.
\end{abstract}

\begin{IEEEImpStatement}
    Simulation-ready digital twins underpin virtual testing in robotics, vehicles, and multi-physics engineering, but authoring correct simulator code often elicits expert-level knowledge. This paper introduces SimBench, a multi-turn benchmark that evaluates large language models on generating and iteratively refining PyChrono digital-twin scripts across multibody, FEA, vehicle, robotics, and sensor scenarios. By combining a curated expert reference dataset with an interpretable rubric-based LLM judge and partial anchoring to execution outcomes, SimBench enables reproducible ranking and diagnostic error analysis beyond surface-level similarity metrics. The benchmark and associated artifacts support rigorous evaluation, guide model improvements for simulator-ready code generation, and provide a template for extending benchmark construction to other simulation ecosystems.
\end{IEEEImpStatement}


\section{Introduction}
\label{into}
% THIS HAS BEEN SAID ELSEWHERE AND IN CONCLUSIONS \textbf{We should say that the proposed rule-based J-LLM benchmark is a much better benchmark than even compile based benchmark compile@1, also much better than similarity based metrics like CodeBLEU and ROUGE-LSUM. The J-LLM benchmark is much closer to the ultimate performance metric pass@1, which is the real performance of the S-LLMs. The J-LLM benchmark is much more efficient than execution based benchmarks, as simulation tasks are expensive to run. A complex simulation task like high-fidelity computational fluid dynamics (CFD) can take months to run. The J-LLM benchmark is a much better way to evaluate the performance of S-LLMs in generating DTs. The J-LLM benchmark uses Reference code and API documentation in the determination process, which may increase the API cost, but the cost of the API is acceptable. We should grow the dataset used to improve our J-LLM, including more dynamic systems, e.g., fluid-solid interaction; multi-phase flow. The rule-based J-LLM can be a better judge compared to other metrics, as it is more closed to the pass@1, which shows the real and ultimate performance of the S-LLMs for simulation tasks.}
% THIS HAS BEEN SAID ELSEWHERE AND IN CONCLUSIONS \textbf{Image, you can know the code by reading even better than by runing the code itself.
%We also should say that we build a really good quality and difficult simulation/DT dataset, although most of LLMs are pretrain on, even the most advanced S-LLMs only have less than $13\%$ pass@1.  }

\IEEEPARstart{I}{t} is only a matter of time before a non-expert will be able to interact with an S-LLM in order to produce a sophisticated digital twin (DT) for a physical system of interest. The DT request might read ``generate a model of a VIPER rover operating on regolith and equipped with two monocular cameras, one stereo camera, and an autonomy stack that uses sensor feeds to ensure obstacle avoidance while operating in lunar conditions.'' Such a model can be generated for a target simulator should an expert dedicate enough time to the task. However, the interest here lies in the scenario where the DT generation falls upon an LLM. Specifically, this contribution proposes a benchmark that measures the proficiency of such an S-LLM when it comes to generating DTs for a target simulator. 

We work under the assumption that a DT assumes the form of a simulator-ready script or a collection of configuration files that enable a virtual experiment via computer simulation. In the VIPER rover example above, the DT is instantiated through hundreds of lines of Python/C++ code or a proprietary modeling language. Alternatively, the DT may be specified via structured configuration files (e.g., JSON), which can likewise span hundreds to thousands of lines. For instance, the rover definition must specify, for each wheel, the center of mass, mass moments of inertia, initial pose and velocity, collision and visualization meshes, and frictional parameters; analogous properties must be defined for the chassis. Camera models require parameters spanning optics and photometry (e.g., aperture, exposure, focal length, BRDF-related quantities, and image-signal-processing settings). The terrain model may involve bulk density, internal friction angle, cohesion, and particle-size distribution. Finally, an autonomy stack may be integrated by configuring a ROS2 environment and selecting perception, planning, and control components. Ultimately, the objective is for an S-LLM to generate the DT construction artifact (e.g., Python code or configuration files); when the user request is underspecified, the S-LLM should either adopt principled defaults or elicit the missing information through targeted clarification.

Analyzing complex multi-physics phenomena via simulation requires sophisticated DTs. 
S-LLMs can potentially handle these through multiple rounds of interaction, 
incorporating natural language feedback from humans. 
To the best of our knowledge, dedicated evaluation protocols for the task of generating DTs are nonexistent, 
as generic similarity-based metrics like CodeBLEU \cite{ren2020codebleu} and ROUGE-L \cite{lin-2004-rouge} 
do not align well with real-world simulation applications and can not capture the complexity of the simulation tasks.

Furthermore, execution-based benchmarks such as Pass@k \cite{chen2021evaluating} can be overly stringent in simulation-centric settings: a single minor defect (e.g., an omitted parameter, setting up an inadequate time step convergence criteria, or swapping value of the time step for that of the relative solution tolerance) may cause a total test failure, yielding a zero score despite the generated digital twin being qualitatively close to correct. More importantly, designing high-coverage and reliable unit tests for complex physics simulations is inherently difficult and rarely scalable. Unlike LeetCode-style problems, where inputs, outputs, and edge cases can be precisely specified, simulation tasks involve long-horizon dynamics, continuous states, numerical sensitivity, and complex scenario logic, making exhaustive, stable, and reusable unit-test suites impractical to author and maintain. As a result, execution-based evaluation becomes costly to extend across diverse systems, scenarios, and multi-turn iterations. It also makes downstream approaches that depend on test-verifiable reward signals (e.g., reinforcement learning with verifiable rewards) substantially less tractable, since they would require scenario-specific test design to provide reliable, dense supervision.

Most importantly, conventional execution-based metrics (e.g., Compile@k and Pass@k)
typically collapse evaluation into a single scalar, 
offering little diagnostic insight into what is wrong and why it is wrong in the DT generated by the S-LLM.
This lack of interpretability is particularly problematic for simulation-grade DTs,
which often comprise hundreds to thousands of lines of code. Indeed, a single scalar score provides limited guidance for debugging, model improvement, or iterative refinement.
In practice, what is needed is an evaluation signal that is both reliable and interpretable, i.e., one that can
attribute errors to concrete aspects of the generated artifact (e.g., missing components,
incorrect parameterization, inconsistent physics assumptions, or incomplete scenario logic),
and provide actionable feedback.

Against this backdrop, the goal of this contribution is to establish an interpretable benchmark for assessing how well
simulation-focused LLMs (S-LLMs) generate DTs. Beyond serving as an evaluation metric,
such an interpretable scoring signal can also support future learning paradigms that require dense and
informative rewards, e.g., improving an S-LLM via training objectives that do not rely on writing
exhaustive unit-test suites for every scenario.
        
Insofar as the nomenclature used in the contribution is concerned, an analysis task in the real world corresponds to a simulation scenario in the digital world. A physical system, e.g., the VIPER rover, has a digital twin. Several digital twins, e.g., those of the rover, terrain, sensor suite, are combined in a simulation scenario towards running a virtual experiment.
 In the absence of a sim-to-real gap, the results collected in simulation are representative of results that
  would have been collected had the experiment taken place in the real world.
   Here, the expectation is that an S-LLM is used to generate the required digital twin[s],
    place them in a simulation scenario, and run a virtual test of interest in a simulator of choice.
     This contribution is not concerned with how a good S-LLM should be set up.
      Rather, the interest is in measuring the ability of an S-LLM to generate good DTs;
       i.e., in benchmarking the S-LLM's performance. As such, the primary interest is in the J-LLM, 
and without loss of generality for the methodology discussed, the simulator used to introduce SimBench is
 PyChrono \cite{pychrono2022}, the Python wrapper of the multi-physics Chrono simulator \cite{chronoOverview2016}. 

The PyChrono choice is justified by the authors' familiarity with the simulator, its broad use in the CAE community, and the fact that it is open source and available on GitHub \cite{projectChronoGithub,projectChronoWebSite}. PyChrono, which was released in 2022 and provides Python bindings to the core C++ Chrono library, has been downloaded from Conda-forge more than 160,000 times \cite{chronodownload}. The Chrono User Forum has more than 800 users \cite{projectChronoForum}. Chrono::Engine provides core functionality for multibody dynamics and nonlinear finite element analysis in the presence of friction and contact. Chrono::Vehicle provides a collection of wheeled and tracked vehicles, facilitating high-fidelity vehicle dynamics analysis, engine and driveline simulation, and deformable terrain modeling. Chrono supports fluid-solid interaction and granular dynamics simulation. Chrono::ROS provides a ROS bridge that enables Chrono to interact with third party autonomy stacks \cite{ROS-2009}. Chrono::Sensor provides sensor modeling and simulations support. Several other modules provide ancillary support for interfacing to SolidWorks, MATLAB, and other third party solvers via FMU/FMI interfacing \cite{FMI_3.0}.

This work makes the following contributions:

(i) \textbf{SimBench: a multi-turn, interpretable benchmark for DT generation.} We introduce SimBench, a benchmark methodology designed to assess an LLM's ability to produce \emph{simulator-ready} digital twins (DTs) within a real simulation ecosystem via multi-turn code editing. Relative to prior code-generation and simulation benchmarks, SimBench explicitly combines (1) multi-turn interaction, (2) rubric-grounded, interpretable scoring with reference/document conditioning, and (3) partial execution anchoring to connect model outputs to runtime behavior. As DT technology becomes increasingly relevant across engineering and scientific domains \cite{digitalTwinReportNAcademies2023}, automating DT construction is highly desirable; a principled, transparent, and interpretable evaluation protocol represents a step toward developing effective simulation-focused LLMs (S-LLMs).

(ii) \textbf{A curated expert dataset and a calibrated J-LLM evaluation pipeline.} We develop a curated dataset authored by simulation experts, consisting of 102 exemplar demonstrations spanning 34 distinct physical systems. For each system, we provide three demonstrations with progressively increasing complexity, together with instructions that guide the transformation from simpler to more intricate configurations. This dataset serves both as (a) a benchmark substrate for standardized evaluation, and (b) a template for scaling data generation for supervised fine-tuning. Using SimBench, we evaluate 33 S-LLMs and collect over 3{,}000 multi-turn dialogues, each paired with expert-assessed scores. This collection can further serve as a preference-style dataset for training or calibrating judge models (J-LLMs) and/or reward models that provide interpretable supervision without requiring scenario-specific unit-test engineering. All datasets, benchmarks, and evaluation code are open-sourced.\footnote{\url{https://github.com/uwsbel/SimBench}}

(iii) \textbf{Generality beyond Chrono.} While SimBench is instantiated for Chrono-based DT generation, the methodology is simulator-agnostic. Given an analogous expert dataset for another simulation platform (e.g., OpenFOAM \cite{openFoam} or PyBullet \cite{bulletPhysicsEngine2020}), the same workflow can be used to construct a calibrated J-LLM suitable for judging DT artifacts in that target ecosystem.

\begin{figure*}[htbp]
    \centering
    \includegraphics[width=0.875\textwidth]{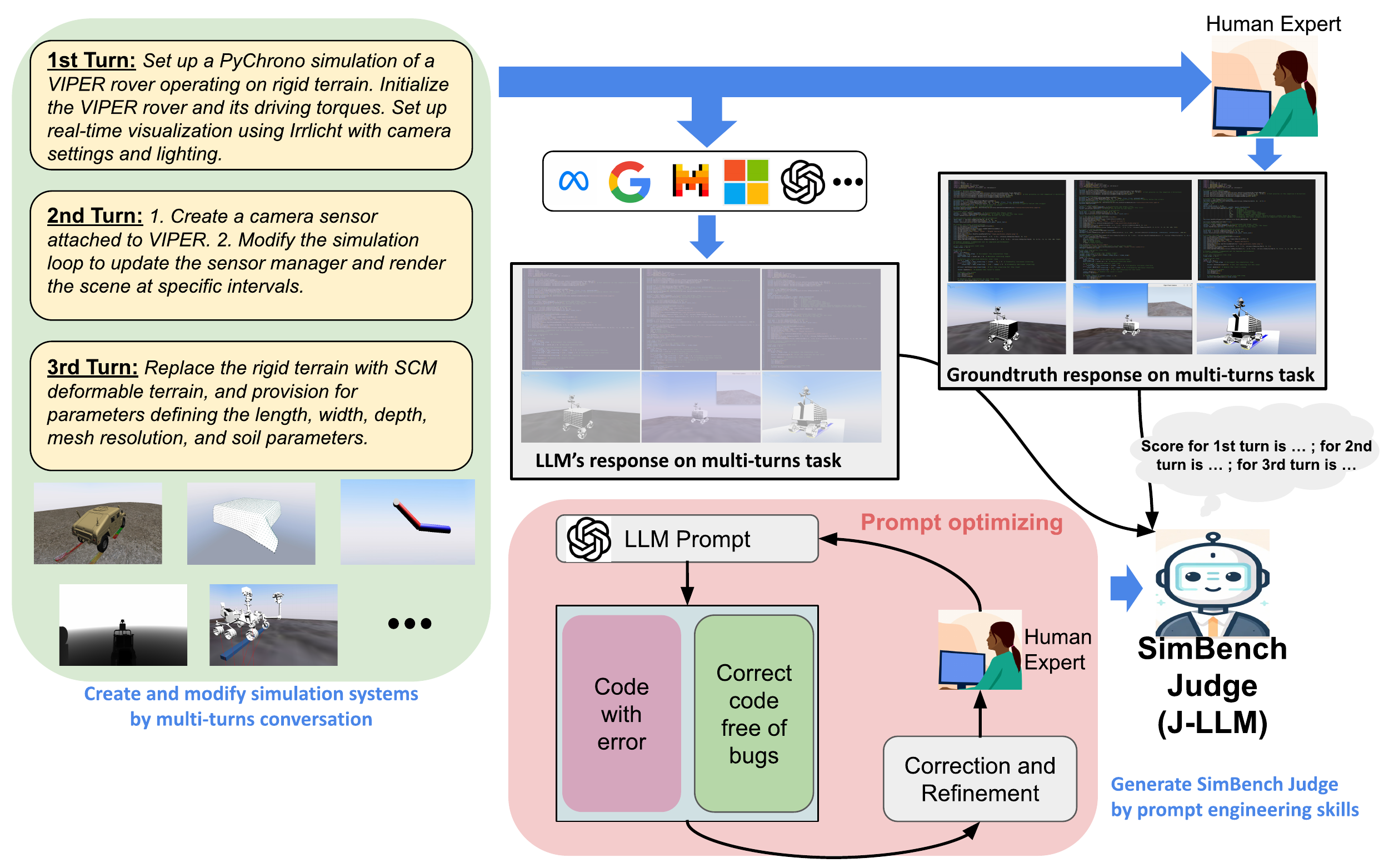}
    \caption{The SimBench pipeline for evaluating S-LLMs. The J-LLM is calibrated using a validation set containing pairs of ground truth and generated DTs. The prompts given to the J-LLM are iteratively optimized to match the score provided by the expert. Then the J-LLM is used to evaluate the S-LLM based on the generated DTs, ground truth DTs, and the API documentation.}
    \label{fig:your-label}
\end{figure*}

\section{Related Work}
\label{sec:related}
\noindent\textbf{LLMs for Digital Twins and Simulation Generation.}
Recent work has begun to explore how LLMs can support digital-twin (DT) and simulation-related workflows.
In \cite{xia2024LLM_multi_agent_simulation}, the authors study LLMs for multi-agent DT simulations in a simplified 2D grid-world setting, and evaluate outputs via similarity between generated code and a reference implementation.
While this provides an initial validation, such similarity-based evaluation can under-specify functional correctness and may not transfer cleanly to simulator-grade DTs with richer physics, longer horizons, and more heterogeneous components.
Complementary efforts investigate LLM-driven decision-making in multi-agent simulations \cite{zhang2024llmDT_decision}, demonstrating the potential for automating decision processes in dynamic environments.
Additionally, \cite{mu2024rule} examines rule-based reward design for training LLMs in controlled settings, motivating the need for evaluation and reward signals that remain reliable as task complexity grows.

\noindent\textbf{Coding Benchmarks and Execution-Based Evaluation.}
A large body of work evaluates code-generation models via unit tests or execution-based correctness.
AlphaCode assesses generated programs against hidden test cases in competitive programming settings \cite{li2022competition}.
CodeT uses self-generated unit tests to score correctness of function implementations \cite{chen2022codet}.
Beyond similarity metrics such as BLEU \cite{Papineni2002BleuAM}, ROUGE \cite{lin-2004-rouge}, and CodeBLEU \cite{ren2020codebleu}, many benchmarks increasingly emphasize execution-based evaluation, where generated code is run and compared against test cases \cite{austin2021program, chen2021evaluating, huang-etal-2022-execution, Lai2022DS1000, hendrycksapps2021}.
Execution-based datasets such as ExeDS \cite{huang-etal-2022-execution} and ODEX \cite{wang2023executionbased} further highlight the importance of testing code in realistic execution environments.
More recently, multi-turn evaluation settings have been proposed to better reflect interactive programming workflows.
For example, MINT evaluates coding LLMs with tool usage and language feedback in multi-turn interactions \cite{wang2024mintevaluatingllmsmultiturn}, and \cite{zhang2024pybenchevaluatingllmagent} benchmarks LLM agents operating in a multi-agent setting with local file-system interactions.

\noindent\textbf{LLM-as-a-Judge.}
There is also growing interest in using LLMs as evaluators for generated outputs.
Prior work studies LLM judges for multi-turn conversations and chain-of-thought reasoning \cite{zheng2023LLMasJudge}.
However, several studies caution that LLM-based evaluation can be biased or unreliable under certain conditions, including when evaluating code or text generation \cite{abdelaziz2021AAAI_benchmark_code_understanding, wang2023llm_unfair_eval}.
Relatedly, \cite{yuan2024selfrewardinglanguagemodels} proposes a self-rewarding paradigm in which an LLM serves as both generator and evaluator, forming a closed-loop refinement process.
These findings motivate careful judge design, calibration, and validation, especially in high-stakes domains such as simulator-ready DT generation.

\noindent\textbf{Positioning of This Work.}
A comprehensive survey of deep learning methods for code generation and evaluation is provided in \cite{wu2022surveydeeplearningmodelscodes}.
In contrast to general-purpose code benchmarks, our focus is on simulator-grade DT construction, where artifacts may span hundreds to thousands of lines and correctness is multi-faceted (physics consistency, component completeness, parameter validity, and scenario logic).
We introduce a benchmark methodology that couples a curated DT-generation dataset with an interpretable, rubric-based judge model (J-LLM) tailored to DT code generation.
This design enables statistically grounded comparisons against similarity-based and execution-based evaluation, while providing diagnostic feedback beyond a single scalar pass/fail outcome.

\section{SimBench Benchmark Design}
\label{sec:simbench}
\IEEEPARstart{S}{imBench} targets simulator-grade digital-twin (DT) generation for Project Chrono/PyChrono, and spans a diverse set of scenarios drawn from multibody systems (MBS), finite element analysis (FEA), vehicle dynamics (VEH), sensor integration (SEN), and robotics (RBT). 
Across these categories, SimBench evaluates an S-LLM’s ability to (i) construct simulator-ready DT artifacts from natural-language requests, and (ii) iteratively refine and extend existing DT implementations in a multi-turn workflow (including targeted modifications and bug fixing). 
In this work, S-LLMs are treated as black-box code generators; developing or selecting S-LLMs is outside the scope. Our focus is the benchmark design and the judge model (J-LLM) used for interpretable evaluation.

\begin{figure}[h]
    \centering
    \includegraphics[width=0.5\textwidth]{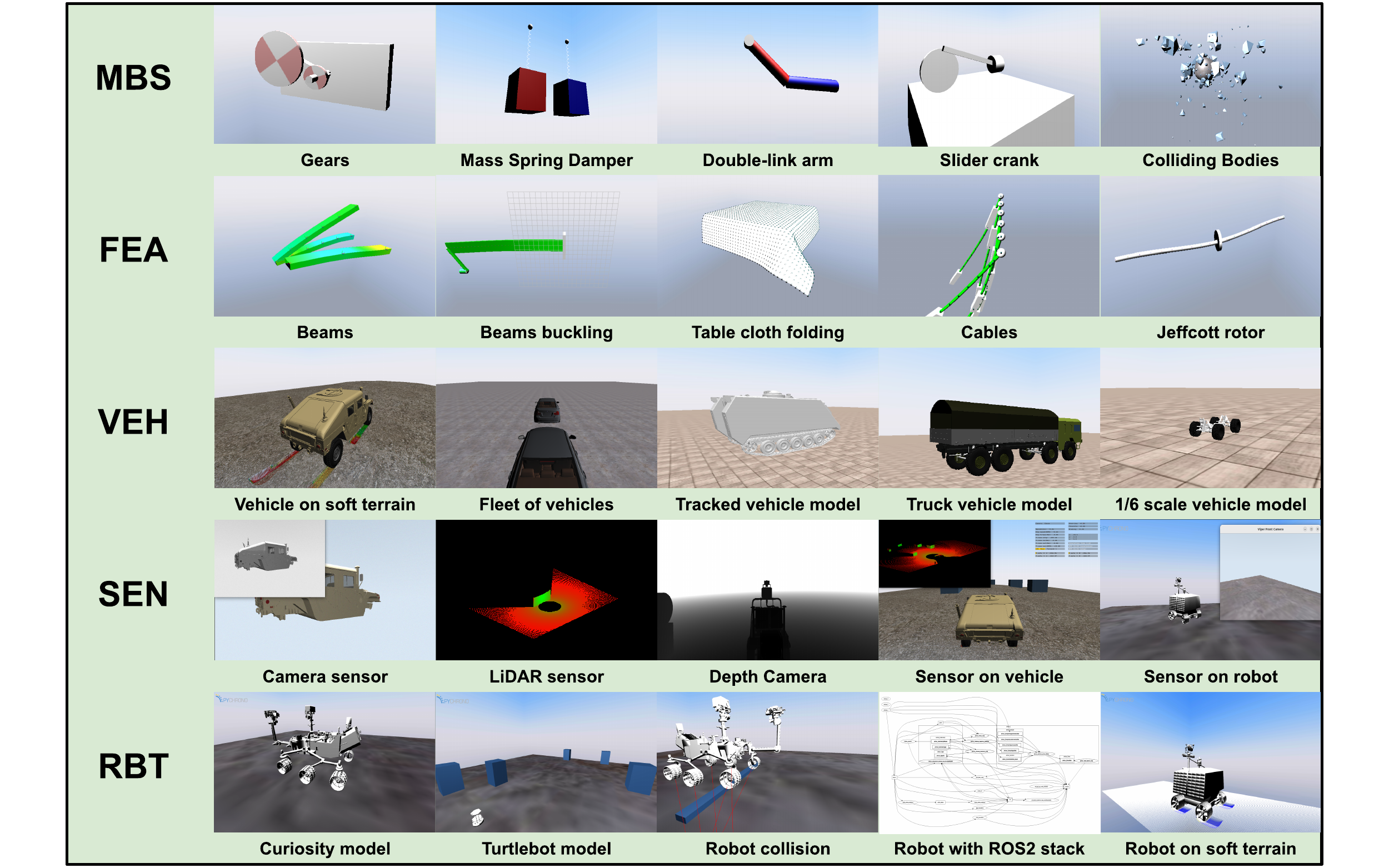}
    \caption{A subset of simulation scenarios in SimBench.}
    \vspace{-10pt}
    \label{fig:demo_gallery}
\end{figure}

\vspace{2pt}
\noindent The families of DTs assessed in SimBench include:
\begin{itemize}
    \item \textbf{Multibody Systems (MBS):} Mechanism and contact-rich systems (e.g., articulated arms, gear trains, slider--crank mechanisms) that exercise rigid-body dynamics, constraints, and contact/friction modeling.
    \item \textbf{Finite Element Analysis (FEA):} Deformable-body scenarios (e.g., cables, beams, shells, plates) that test proficiency in discretization choices, boundary conditions, loading, and solver configuration.
    \item \textbf{Vehicle Dynamics (VEH):} On-road and off-road vehicles (e.g., city bus, HMMWV, M113, Kraz, MAN, sedans) with driver/powertrain/tire models and scenario logic (e.g., lane changes, logging, mixed traffic).
    \item \textbf{Sensor Integration (SEN):} GPS/IMU/LiDAR/camera configurations that evaluate correct sensor instantiation, attachment, calibration, and data acquisition pipelines.
    \item \textbf{Robotics (RBT):} Robotic platforms (e.g., Turtlebot, Curiosity, VIPER) and off-road interactions such as deformable terrain and soil contact models (SCM) \cite{Krenn2011}.
\end{itemize}
SimBench draws on $N_{\text{sys}}=34$ expert-authored physical systems, each realized as a three-turn demonstration, yielding $N_{\text{task}}=102$ turn-level benchmark tasks across these categories. Each system is represented by three high-quality turns with progressively increasing complexity, enabling robust assessment of code generation and code modification capabilities. Figure~\ref{fig:demo_gallery} illustrates representative SimBench scenarios.

\medskip

{\noindentBold{Task Formulation and Multi-Turn Protocol}}: 
SimBench tasks are expressed as natural-language requests that vary in specificity. We distinguish two request styles:

\begin{itemize}
    \item \textbf{Vague Requests:} high-level instructions that intentionally leave parts of the DT specification underspecified (e.g., missing parameter values or implementation details). These prompts test whether an S-LLM can infer reasonable defaults, make meaningful modeling choices, and produce a coherent simulator-ready DT under incomplete information.
    \item \textbf{Sharp Requests:} explicit, constraint-rich instructions that precisely specify the desired DT changes and expected behaviors. These prompts test faithful instruction following and targeted code editing in response to concrete requirements.
\end{itemize}

Each SimBench task is structured as a \emph{three-turn} interaction that reflects a common DT-development workflow: an initial high-level request followed by increasingly specific refinements. In our benchmark instantiation, the first turn is \emph{typically} a Vague Request, while the second and third turns are Sharp Requests.
Table~\ref{tab:mt_bench_sample} provides representative multi-turn prompts across categories.

\begin{enumerate}
    \item \textbf{Turn 1 (Vague Request): DT construction from scratch.}
    The first turn requests building a DT from an abstract or minimally specified description (e.g., a basic mechanism, a deformable object setup, or a vehicle/robot scenario). The goal is to probe foundational knowledge of PyChrono primitives and the ability to make principled modeling decisions when details are missing.

    \item \textbf{Turn 2 (Sharp Request): targeted modification and bug fixing.}
    The second turn provides concrete requirements for modifying the previously generated DT (e.g., replacing a joint type, adding a sensor, changing initial conditions, or adjusting solver/visualization settings). In addition, SimBench monitors \emph{bug-fixing} requests where a DT implementation is intentionally perturbed (e.g., API misuse, inconsistent parameters, or missing configuration), and the S-LLM is instructed to identify and correct the code.

    \item \textbf{Turn 3 (Sharp Request): challenging extension of the scenario.}
    The final turn specifies a more demanding extension that typically requires multi-step reasoning and coordinated changes across components (e.g., introducing autonomy/control logic, executing a lane-change scenario, integrating multiple sensors, adding additional agents/objects, or logging task-relevant state). This turn evaluates an S-LLM’s ability to maintain consistency with prior turns while scaling the DT to a more complex virtual experiment.
\end{enumerate}

\begin{table}[h!]
	\centering
	\begin{tabular}{|c|c|c|c|}
		\hline
		\multirow{2}{0.7cm}{Genre} & \multicolumn{3}{c|}{Multi-turn Task Description}  \\ \cline{2-4}
		& 1st turn & 2nd turn & 3rd turn   \\ \hline
		\parbox[c][1.75cm][c]{0.7cm}{MBS} & \parbox[c][1.75cm][c]{1.5cm}{Generate a one-link arm} & \parbox[c][1.75cm][c]{2.1cm}{Replace revolute joint with spherical joint} & \parbox[c][1.75cm][c]{2.1cm}{Extend single link into double-link arm}  \\ \hline
		\parbox[c][2cm][c]{0.7cm}{FEA} & \parbox[c][2cm][c]{1.5cm}{Model the folding of tablecloth} & \parbox[c][2cm][c]{2.1cm}{Change viz. \& solver \& apply load; set boundary conditions} & \parbox[c][2cm][c]{2.1cm}{Enhance to higher quality visualization}  \\ \hline
		\parbox[c][2.5cm][c]{0.7cm}{VEH} & \parbox[c][2.5cm][c]{1.5cm}{Create Kraz truck vehicle model} & \parbox[c][2.5cm][c]{2.1cm}{Adjust truck initial position and viz. angle; implement double lane change} & \parbox[c][2.5cm][c]{2.1cm}{Switch scene to highway; add another sedan; record truck state info}  \\ \hline
		\parbox[c][2cm][c]{0.7cm}{SEN} & \parbox[c][2cm][c]{1.5cm}{Generate LiDAR to sense obj. defined via mesh} & \parbox[c][2cm][c]{2.1cm}{Replace mesh with box object and add one more 2D LiDAR sensor} & \parbox[c][2cm][c]{2.1cm}{Create LiDAR and camera sensors attaching to a scaled vehicle}  \\ \hline
		\parbox[c][1.75cm][c]{0.7cm}{RBT} & \parbox[c][1.75cm][c]{1.5cm}{Generate a Turtlebot robot} & \parbox[c][1.75cm][c]{2.1cm}{Showcase robot moving right, left, and straight ahead} & \parbox[c][1.75cm][c]{2.1cm}{Add 5 boxes randomly \& use LiDAR to monitor scene}  \\ \hline
	\end{tabular}
	\caption{Sample task requests submitted to the S-LLM.}
	\vspace{-10pt}
	\label{tab:mt_bench_sample}
\end{table}

\medskip
\noindent\textbf{Example: Curiosity rover (RBT) multi-turn DT generation (abridged).}
\label{app:curiosity_example}
We provide an illustrative robotics (RBT) task to demonstrate the three-turn protocol.
Turn~1 is a \emph{vague} request that specifies the overall DT goal, while Turns~2--3 are \emph{sharp} requests that require targeted code edits and extensions.
In SimBench, Turns~2--3 additionally include the full code context from the previous turn; we omit the code context here for readability.

\begin{PromptBlock}{Curiosity (RBT) --- Turn 1 (Vague Request; DT construction)}
    Create a PyChrono simulation showcasing a Curiosity rover navigating on a rigid terrain.
    The system includes a fixed ground with collision properties and a rover model equipped with a motor control driver for real-time steering inputs.
    Visualize the simulation using the Irrlicht visualization system with customizable camera settings, lighting, shadows, and additional visual elements like textures and logos.
    \end{PromptBlock}
    
    \begin{PromptBlock}{Curiosity (RBT) --- Turn 2 (Sharp Request; code editing)}
    Do the following changes based on the provided code:
    (*@\textbf{1.}@*) Add a long box obstacle for the rover to cross.\\
    (*@\textbf{2.}@*) Change the initial position of the rover from $(0,\,0.2,\,0)$ to $(-5,\,0.0,\,0)$.\\
    (*@\textbf{3.}@*) Set the rover to move forward with zero steering input, instead of gradually increasing steering input over time.
    \end{PromptBlock}
    
    \begin{PromptBlock}{Curiosity (RBT) --- Turn 3 (Sharp Request; sensor integration)}
    Do the following changes to the given code:
    (*@\textbf{1.}@*) Add the sensor module (\texttt{pychrono.sensor}) import.\\
    (*@\textbf{2.}@*) Create a sensor manager (\texttt{sens.ChSensorManager(system)}).\\
    (*@\textbf{3.}@*) Add a LiDAR sensor attached to the rover chassis:
    \begin{itemize}
      \item Define LiDAR parameters such as update rate, horizontal and vertical samples, and field of view.
      \item Create the LiDAR sensor (\texttt{sens.ChLidarSensor}) and attach it to the rover chassis.
      \item Set up filters for the LiDAR data.
      \item Add the LiDAR sensor to the sensor manager.
    \end{itemize}
    (*@\textbf{4.}@*) Update the sensor manager in the simulation loop (\texttt{manager.Update()}).
    \end{PromptBlock}

% Optional (recommended): one sentence to reinforce "interpretability" link
\noindent This example also motivates the need for interpretable evaluation: the sharp requests in Turns~2--3 require localized, component-level modifications (obstacle geometry, initial conditions, driver logic, and sensor pipeline), which are difficult to diagnose using a single pass/fail outcome.

\medskip

{\noindentBold{Evaluation framework and modalities}:}
\label{sec:modalities} In the approach proposed, a rubric-based J-LLM evaluates the performance of an S-LLM using the tasks contained in the SimBench environment. Unlike traditional metrics such as BLEU, ROUGE, and CodeBLEU, the J-LLM is designed to provide a comprehensive evaluation of the S-LLM's performance in multi-turn simulation tasks. The J-LLM is instructed via a rubric-driven prompt that provides (i) the candidate code, 
(ii) an expert reference solution (when available), and optionally (iii) a compact API documentation excerpt.
The prompt elicits both a scalar score and itemized, component-level deductions.

Three evaluation modalities are considered in SimBench.
In \textbf{J-LLM\_Ref\_Doc}, the judge is prompted with both the expert reference code and an expert-written API documentation excerpt (approximately 4{,}000 tokens), in addition to the S-LLM’s candidate code.
This setting requires a long context window to jointly accommodate documentation, reference, and candidate code; in our experiments, this constraint limits the applicability of judge models with shorter context windows.

In \textbf{J-LLM\_Ref}, the judge is provided the expert reference code and the candidate code, without additional documentation, and must rely on general coding priors together with the reference implementation.

Finally, in \textbf{J-LLM\_Doc}, the judge is provided only the expert-written API documentation excerpt and the candidate code, without access to the reference implementation

\medskip

{\noindentBold{Execution-Based Evaluation}}: In addition to J-LLM evaluation, SimBench incorporates execution-based metrics that provide ground-truth validation of generated code:

\begin{itemize}
    \item \textbf{Compile@1}: Measures whether the generated code is parsed successfully in Python. This metric captures the S-LLM's ability to produce syntactically correct PyChrono code. For simulators other than PyChrono, this test might amount to a C or C++ compilation.
    \item 
    \item \textbf{Pass@1}: Measures whether the generated code passes predefined functional tests. Unlike Compile@1, Pass@1 requires the code to be semantically correct and to produce the expected simulation behavior, as evidenced by a successful pass, correct numerical outputs, and physically reasonable visualizations.
\end{itemize}

\medskip

{\noindentBold{Code Similarity Metrics}}: SimBench also computes automated code similarity metrics:

\begin{itemize}
    \item \textbf{CodeBLEU}~\cite{ren2020codebleu}: A composite metric that considers n-gram match, weighted n-gram match, syntax match (AST), and dataflow match. CodeBLEU captures both surface-level and structural code similarity.
    \item \textbf{ROUGE-L}: Measures the longest common subsequence between generated and reference code, capturing sequential similarity.
\end{itemize}

\medskip
\noindent\textbf{Rubric-prompt design for interpretable judging.}
SimBench adopts a rubric-driven judge prompting strategy to obtain both a scalar score and localized, component-level feedback.
The J-LLM is prompted with (i) the candidate DT code produced by an S-LLM, (ii) an expert reference implementation for the same turn, and optionally (iii) PyChrono API documentation.
The prompt enforces a fixed, additive deduction scheme across six categories (Completeness, Correctness, Code Quality, Efficiency, Robustness, and Visualization), with explicit point-deduction guidelines.
This structure encourages the judge to (a) identify \emph{what} is missing or incorrect, (b) explain \emph{why} it matters relative to the reference or best practices, and (c) translate that diagnosis into a consistent numeric score.

\vspace{2pt}
\noindent
The rubric categories are intentionally aligned with common failure modes in simulator-grade DT code generation.
For example, structural omissions (e.g., missing sensor manager updates, missing collision settings, absent visualization blocks) map naturally to \emph{Completeness}; API misuse and frame/pose inconsistencies map to \emph{Correctness}; and brittle or non-idiomatic implementations map to \emph{Robustness} and \emph{Code Quality}.
To mitigate known LLM-judge biases, the prompt additionally includes guardrails against position bias, length bias, and stylistic preferences. It also requires the final score to be emitted in a machine-parsable format \texttt{[[x]]} for downstream aggregation.

\vspace{2pt}
\noindent\textbf{Judge calibration protocol.}
To calibrate the rubric-grounded judge prompt, we constructed a calibration set covering the major Chrono task classes. Concretely, we selected one representative task from each of five classes (\texttt{MBS}, \texttt{FEA}, \texttt{SEN}, \texttt{RBT}, \texttt{VEH}), yielding $N_{\text{ref}}=5$ expert reference scripts. We use one task per class because tasks within a class share core APIs, structure, and recurring failure modes, making a representative reference sufficient for prompt calibration without overfitting to a narrow instance.

For each reference script, we manually synthesized $N_{\text{var}}=5$ perturbed variants by injecting common implementation errors frequently discussed in the Project Chrono community forum \cite{projectChronoForum}, while keeping the underlying task intent unchanged. The injected bugs span: (i) API hallucination or misuse (incorrect class/method names, deprecated calls, invalid signatures), (ii) constraint/joint specification errors (missing constraints, incorrect frames, inconsistent coordinates), (iii) simulation-loop and subsystem omissions (e.g., missing sensor-manager updates, missing module initialization, incorrect update order), (iv) integrator/solver misconfiguration (inappropriate system type, solver choice, time step, or tolerances), and (v) visualization/IO omissions (missing visualization setup, assets, render/update calls, or output logging).

We iteratively refined the judge prompt using only this calibration set until (a) the induced ranking of a small panel of S-LLMs agreed with expert rankings, and (b) no systematic judge failure modes were observed on a small holdout subset of perturbed cases.

\medskip

\noindent\textbf{J-LLM scoring, aggregation, and reporting.}
SimBench contains $N_{\text{sys}}=34$ distinct physical systems. Each system is evaluated through a fixed three-turn protocol ($t\in\{1,2,3\}$), yielding $N_{\text{task}}=3N_{\text{sys}}=102$ turn-level benchmark tasks.

For a given S-LLM, let $y_{i,t}$ denote the code produced for system $i$ at turn $t$. Under a chosen evaluation modality (Sec.~\ref{sec:modalities}), the judge model (J-LLM) is \emph{prompted} to output (i) an interpretable textual critique with itemized deductions and (ii) a scalar score:
\begin{equation}
s_{i,t} \;\triangleq\; \mathrm{JLLM}\!\left(y_{i,t};\, r_{i,t}, d\right)\in[0,100],
\end{equation}
where $r_{i,t}$ is the expert reference implementation when available, and $d$ denotes the optional API-documentation context.

\vspace{2pt}
\noindent\emph{System-level aggregation.}
We first aggregate the three turn scores into a system-level score:
\begin{equation}
S_i \;\triangleq\; \sum_{t=1}^{3} w_t\, s_{i,t},
\qquad \text{with}\quad \sum_{t=1}^{3} w_t = 1,
\label{eq:sys_agg}
\end{equation}
using uniform weights $w_1=w_2=w_3=\tfrac{1}{3}$ to reflect end-to-end workflow competence.
\vspace{2pt}

\noindent\emph{Overall score.}
We report the overall SimBench score as the task-level macro-average over all $N_{\text{task}}=102$ turn-level tasks:
\begin{equation}
S \;\triangleq\; \frac{1}{N_{\text{task}}}\sum_{i=1}^{N_{\text{sys}}}\sum_{t=1}^{3} s_{i,t}.
\label{eq:overall_agg}
\end{equation}
Equivalently, since each system contributes exactly three turns, $S = \frac{1}{N_{\text{sys}}}\sum_{i=1}^{N_{\text{sys}}} S_i$.

\vspace{2pt}
\noindent\emph{Category-level reporting.}
Let $\mathcal{T}_c$ denote the set of \emph{turn-level} tasks $(i,t)$ that belong to category $c$. We report category scores as
\begin{equation}
S^{(c)} \;\triangleq\; \frac{1}{|\mathcal{T}_c|}\sum_{(i,t)\in\mathcal{T}_c} s_{i,t}.
\label{eq:cat_agg}
\end{equation}

{\noindent\textbf{Dataset Complexity Analysis}}: To contextualize SimBench's complexity, we provide a quantitative comparison with established code generation benchmarks in Table~\ref{tab:benchmark_comparison}. Token counts are computed using OpenAI's \texttt{tiktoken} library with the \texttt{cl100k\_base} encoding.

\begin{table}[h!]
    \centering
    \caption{Comparison of SimBench with existing code generation benchmarks. SimBench exhibits significantly higher complexity in both prompt and solution lengths.}
    \label{tab:benchmark_comparison}
    \begin{tabular}{l r r r r r}
        \toprule
        \textbf{Benchmark} & \textbf{Tasks} & \textbf{Prompt} & \textbf{Solution} & \multicolumn{2}{c}{\textbf{Ratio to SimBench}} \\
        & & \textbf{(tokens)} & \textbf{(tokens)} & Prompt & Solution \\
        \midrule
        MBPP~\cite{austin2021program} & 500 & 16 & 58 & 61.6$\times$ & 24.4$\times$ \\
        MBPP+~\cite{liu2024your} & 378 & 19 & 40 & 51.8$\times$ & 35.4$\times$ \\
        DS-1000~\cite{lai2023ds} & 1,000 & 282 & 42 & 3.5$\times$ & 33.7$\times$ \\
        HumanEval~\cite{chen2021evaluating} & 164 & 131 & 54 & 7.5$\times$ & 26.2$\times$ \\
        HumanEval+~\cite{liu2024your} & 164 & 131 & 54 & 7.5$\times$ & 26.2$\times$ \\
        BigCodeBench~\cite{zhuo2024bigcodebench} & 1,140 & 145 & 112 & 6.8$\times$ & 12.6$\times$ \\
        CodeContests~\cite{li2022competition} & 165 & 593 & 762 & 1.7$\times$ & 1.9$\times$ \\
        \midrule
        \textbf{SimBench (ours)} & \textbf{102} & \textbf{985} & \textbf{1,415} & 1.0$\times$ & 1.0$\times$ \\
        \bottomrule
    \end{tabular}
\end{table}

A key distinction of SimBench lies in its multi-turn structure. Unlike single-turn benchmarks where each task is independent, SimBench's second and third turns require the S-LLM to modify existing code from the previous turn. This design reflects real-world software development workflows where developers iteratively refine and extend existing codebases. Table~\ref{tab:turn_complexity} details the prompt composition across turns.

\begin{table}[h!]
    \centering
    \caption{Prompt complexity across SimBench turns. Turns 2 and 3 include full code context from previous turns, reflecting realistic code modification scenarios.}
    \label{tab:turn_complexity}
    \begin{tabular}{l r r r r}
        \toprule
        \textbf{Turn} & \textbf{Text} & \textbf{Code Context} & \textbf{Total Prompt} & \textbf{Solution} \\
        & \textbf{(tokens)} & \textbf{(tokens)} & \textbf{(tokens)} & \textbf{(tokens)} \\
        \midrule
        Turn 1 & 87 & --- & 87 & 1,252 \\
        Turn 2 & 189 & 1,252 & 1,441 & 1,416 \\
        Turn 3 & 123 & 1,306 & 1,429 & 1,577 \\
        \midrule
        \textbf{Average} & \textbf{133} & \textbf{853} & \textbf{985} & \textbf{1,415} \\
        \bottomrule
    \end{tabular}
\end{table}

The multi-turn paradigm introduces several challenges absent in conventional benchmarks:
\begin{itemize}
    \item \textbf{Long-context reasoning}: With prompts averaging 985 tokens (and exceeding 1,400 tokens in later turns), S-LLMs must maintain coherence across extended contexts while understanding the existing codebase.
    \item \textbf{Code comprehension}: Unlike generation-only tasks, turns 2 and 3 require the S-LLM to first understand the provided code before making targeted modifications, testing both reading and writing capabilities.
    \item \textbf{Incremental complexity}: Solutions grow progressively longer (1,252 $\rightarrow$ 1,416 $\rightarrow$ 1,577 tokens), requiring the S-LLM to manage increasing code complexity while maintaining consistency with prior implementations.
\end{itemize}

Table~\ref{tab:category_complexity} provides a breakdown of complexity by simulation category, revealing that Sensor (SEN) and Finite Element Analysis (FEA) scenarios demand the most extensive code generation.

\begin{table}[h!]
    \centering
    \caption{Complexity by simulation category.}
    \label{tab:category_complexity}
    \begin{tabular}{l r r r}
        \toprule
        \textbf{Category} & \textbf{Systems} & \textbf{Avg. Prompt} & \textbf{Avg. Solution} \\
        & & \textbf{(tokens)} & \textbf{(tokens)} \\
        \midrule
        SEN & 4 & 1,296 & 1,932 \\
        FEA & 5 & 1,172 & 1,598 \\
        MBS & 5 & 1,011 & 1,415 \\
        VEH & 14 & 885 & 1,312 \\
        RBT & 6 & 640 & 922 \\
        \bottomrule
    \end{tabular}
\end{table}

\noindent\textbf{Reproducibility of prompts and qualitative examples.}
All wrapper prompts used to query S-LLMs, as well as the rubric prompts used by the J-LLM, are provided in Appendix~\ref{app:prompt_templates} and Appendix~\ref{app:jllm_prompts}.
A complete three-turn Curiosity rover case study (including expert reference solutions, turn-to-turn patches, and an example S-LLM output with J-LLM feedback) is included in Appendix~\ref{app:curiosity_gt} and Appendix~\ref{app:claude_curiosity_turn3}.

\bigskip

\noindent\textbf{Generality and Portability to Other Simulators.}
Although SimBench is instantiated in this work for Chrono/PyChrono, the benchmark methodology is intended to be \emph{simulator-agnostic}. In particular, SimBench separates (i) the \emph{task protocol} used to elicit multi-turn digital-twin (DT) development behavior from an S-LLM, (ii) the \emph{context channels} used to condition evaluation (expert references and documentation), and (iii) the \emph{execution oracle} used to anchor correctness. Porting SimBench to another simulator therefore amounts to providing simulator-specific implementations of the following three interface points.

\textbf{(1) Task specification (multi-turn protocol and artifact expectations).}
SimBench tasks are defined as a fixed-turn interaction that reflects realistic DT authoring: an initial DT creation request (Turn 1), followed by successive DT modification/extension requests (Turns 2--3) that require understanding and editing the existing artifact. This protocol is independent of the simulator itself; however, the \emph{artifact type} and \emph{acceptance criteria} must be specified per target platform (e.g., Python scripts, input decks, XML/URDF/SDF configurations, plugin code, or mixed pipelines). The same prompt-genres used in SimBench (e.g., \emph{vague} vs.\ \emph{sharp} requests) can be retained to probe different aspects of instruction following, default inference, and code-editing competence.

\textbf{(2) Context sources (documentation and expert references).}
SimBench evaluates generated DTs under optional context channels: (i) simulator/API documentation and (ii) expert reference implementations when available. To port the benchmark, one should provide analogous context for the target simulator: stable API/documentation excerpts that cover the exercised features, and a small expert-authored dataset of reference DT solutions and turn-to-turn patches for the selected scenario families. These references serve two roles: they provide grounding for evaluation and they enable calibration of the rubric-based judge model (J-LLM) to match expert scoring behavior in the target ecosystem.

\textbf{(3) Execution oracle (what ``compile'' and ``pass'' mean).}
Finally, SimBench anchors evaluation using execution-based signals (e.g., Compile@1 and Pass@1), while complementing them with interpretable rubric-based judging. For a new simulator, the benchmark designer must define the platform-appropriate execution oracle, including: (i) what constitutes a syntactic validity check (``compile''), (ii) what constitutes a semantic correctness check (``pass''), and (iii) tolerances, timeouts, and determinism controls. Importantly, the oracle need not mirror PyChrono's toolchain; it must only provide objective, automated checks that capture whether the generated DT artifact is runnable and functionally correct for the intended scenario.

\textbf{Examples.}
To illustrate how these interface points specialize across platforms:
(i) For \emph{Abaqus} \cite{ABAQUSreference}, the ``compile'' analogue is typically an \texttt{abaqus job=} submission that completes without preprocessing/runtime errors (input-deck validity, successful analysis completion), while ``pass'' can be defined via regression checks on output database (ODB) quantities such as displacements/forces/energies, convergence status, and invariants under prescribed loading and boundary conditions.
(ii) For \emph{MuJoCo} \cite{todorovMujoco2012}, the syntactic check corresponds to successful model parsing/loading (XML validity, asset resolution, no warnings/errors that prevent simulation), while ``pass'' can be defined via rollout-based checks (no NaNs/instabilities, constraint satisfaction within tolerance) and task-specific quantitative comparisons (trajectory tracking error, contact event timing, sensor outputs) against reference rollouts.
(iii) For other ecosystems (e.g., OpenFOAM \cite{openFoam}, PyBullet \cite{Pybullet2023}, Isaac Sim \cite{IssacNvidia2023}, Gazebo \cite{gazebo}), the same pattern applies: the task protocol and rubric remain stable, while the artifact formats, documentation corpus, and execution oracle are adapted to the platform's compilation/loading and correctness notions.

From a high vantage point, Chrono serves as an end-to-end \emph{existence proof} for constructing a calibrated, interpretable, execution-anchored benchmark for simulator-grade DT generation. The methodology transfers to other simulators by re-instantiating the three interface points above: task specification, context sources, and execution oracle.

\section{Benchmarking Experiments}
\label{sec:exp}
\textbf{S-LLMs Evaluated}. SimBench is demonstrated in conjunction with benchmarking a set of 33 S-LLMs,
encompassing both open- and closed-source models of various sizes, architectures, and family.
For closed-source models, we include commercial S-LLMs, such as \textbf{gpt-4.1-nano} and \textbf{o4-mini}
from OpenAI, as well as \textbf{Claude 4.0 Sonnet} from Anthropic, and \textbf{Gemini 2.5pro} from Google.
The open-source models evaluated include from the \textbf{LLaMA-3.1} family (8B, 70B, 405B) to the \textbf{LLaMA-4} family (400B, 109B).
We included specialized models like \textbf{codellama-70B}, \textbf{codestral-22B}, \textbf{CodeGemma-7B}, and also the mamba class \textbf{mamba-codestral-7b} model,
fine-tuned for code generation. We also considered S-LLMs that have been trained on a mix of general and 
domain-specific data, e.g., \textbf{Gemma-2} (2B, 9B, 27B) and \textbf{Mistral} (12b, 8x7b, 8x22B). 
Additionally, models such as \textbf{Phi-3} (3.8B, 7B, 14B) and \textbf{Mixtral} (8x7B, 8x22B) 
are included to explore the impact of different scales and training regimes on their ability to perform in 
multi-turn setups. 

In terms of generation settings, unless otherwise specified, we sample S-LLM outputs with temperature $=0.6$ and top-$p=0.9$ using a fixed prompt wrapper shared across models. We exclude base (non-instruction-tuned) code models to focus on instruction-following behavior. For the judge (J-LLM), we use temperature $=0.2$ and top-$p=0.7$ to reduce scoring variance. The judge prompt is calibrated once and then frozen for all models and tasks. All raw data for evaluation of models under three types of metrics (similarity, J-LLM, and execution-based) are included in the repository.

\subsection{Historical Trends in S-LLM Performance on SimBench}

To evaluate whether S-LLM simulation code generation performance has improved across successive six-month intervals, we analyze the relationship between model release dates and J-LLM scores. Figure~\ref{fig:temporal_trend} plots the average of the three modalities of J-LLM scores (J-LLM\_Ref\_Doc, J-LLM\_Ref, and J-LLM\_Doc) against release dates for all 33 evaluated models. 
A clear positive trend emerges: models released more recently demonstrate stronger performance on SimBench tasks. 

Computing correlation coefficient yields $\rho = 0.624$ ($p < 0.001$), indicating a significant temporal improvement in simulation code generation capabilities. Several notable observations emerge from this temporal analysis. First, the variance in performance has increased over time: while early models (2023 Q4-2024 Q3) cluster in the 30-40 score range, recent models (2025 Q1-Q3) span from 40 to 49, suggesting both continued improvement of frontier models and diversification of capabilities across different model families. Second, reasoning-enhanced models released in 2025 Q1-Q2 consistently occupy the upper performance tier, with Claude-4-Sonnet, o3, and Claude-3.7-Sonnet achieving the top three positions. Third, model scale alone does not guarantee superior performance. Indeed, several smaller recent models (e.g., gpt-4.1-nano, gpt-4.1-mini) outperform earlier large-scale models (e.g., Nemotron-4-340B, Llama-3.1-405B), highlighting the importance of architectural innovations and training methodologies beyond parameter count.

The temporal trend analysis indicates that SimBench is sensitive to meaningful differences in model capability and that the field is making measurable progress in simulation-focused code generation. This trajectory suggests that continued advances, especially improved long-horizon code editing, stronger tool use, and better simulator-specific grounding, should translate into higher performance on complex multi-turn simulation tasks.

\begin{figure*}[htbp]
\centering
\includegraphics[width=\textwidth]{images/jllm_vs_exact_dates.png}
\caption{Temporal evolution of S-LLM performance on SimBench. Y-axis is the average of three modalities of J-LLM scores, they are plotted against model release dates, 
showing a positive trend ($\rho = 0.624$, $p < 0.001$).}
\label{fig:temporal_trend}
\end{figure*}

\begin{table*}[htbp]
  \centering
  \resizebox{\textwidth}{!}{%
  \begin{tabular}{lccccc ccc}
  \toprule
  \textbf{Model} & \textbf{Company} & \textbf{Open Weights} & \textbf{Size} & \textbf{Release} & \textbf{Reasoning} & \textbf{J-LLM-Ref-Doc} & \textbf{J-LLM-Ref} & \textbf{J-LLM-Doc} \\
  \midrule
  claude-4-sonnet & Anthropic & -- & -- & 2025-Q2 & \checkmark & 49 & 39 & 55 \\
  o3 & OpenAI & -- & -- & 2025-Q2 & \checkmark & 46 & 37 & 67 \\
  claude-3-7-sonnet & Anthropic & -- & -- & 2025-Q1 & \checkmark & 43 & 36 & 55 \\
  o4-mini & OpenAI & -- & -- & 2025-Q2 & \checkmark & 42 & 35 & 54 \\
  qwen3-235b-a22b & Alibaba & \checkmark & 235B & 2025-Q2 & \checkmark & 42 & 32 & 55 \\
  Gemini-2.5-pro & Google & -- & -- & 2025-Q2 & \checkmark & 41 & 31 & 62 \\
  gpt-4.1-mini & OpenAI & -- & -- & 2025-Q2 & -- & 41 & 34 & 55 \\
  gpt-4o-mini & OpenAI & -- & -- & 2024-Q3 & -- & 41 & 28 & 49 \\
  llama4-maverick & Meta & \checkmark & 400B & 2025-Q2 & -- & 41 & 33 & 44 \\
  llama4-scout & Meta & \checkmark & 109B & 2025-Q2 & -- & 41 & 34 & 38 \\
  llama-3.3-70b & Meta & \checkmark & 70B & 2024-Q4 & -- & 41 & 34 & 42 \\
  deepseek-r1-32b & DeepSeek & \checkmark & 32B & 2025-Q1 & \checkmark & 40 & 32 & 54 \\
  gpt-4.1-nano & OpenAI & -- & -- & 2025-Q2 & -- & 40 & 34 & 55 \\
  llama-3.1-70b & Meta & \checkmark & 70B & 2024-Q3 & -- & 40 & 33 & 39 \\
  gpt-4.1 & OpenAI & -- & -- & 2025-Q2 & -- & 39 & 34 & 55 \\
  Gemini-1.5-pro & Google & -- & -- & 2024-Q1 & -- & 39 & 33 & 43 \\
  codestral-22b-v0.1 & Mistral & \checkmark & 22B & 2024-Q2 & -- & 39 & 32 & 40 \\
  llama-3.1-405b & Meta & \checkmark & 405B & 2024-Q3 & -- & 39 & 33 & 38 \\
  mixtral-8x22b-v0.1 & Mistral & \checkmark & 176B (8$\times$22B) & 2024-Q2 & -- & 39 & 34 & 37 \\
  llama-3.1-8b & Meta & \checkmark & 8B & 2024-Q3 & -- & 38 & 31 & 35 \\
  mistral-nemo-12b & Mistral & \checkmark & 12B & 2024-Q3 & -- & 38 & 30 & 32 \\
  gemma-2-27b & Google & \checkmark & 27B & 2024-Q2 & -- & 36 & 29 & 37 \\
  mixtral-8x7b-v0.1 & Mistral & \checkmark & 46.7B (8$\times$7B) & 2023-Q4 & -- & 36 & 34 & 31 \\
  claude-3-5-sonnet & Anthropic & -- & -- & 2024-Q2 & -- & 34 & 26 & 53 \\
  deepseek-r1-8b & DeepSeek & \checkmark & 8B & 2025-Q1 & \checkmark & 34 & 32 & 54 \\
  gpt-4o & OpenAI & -- & -- & 2024-Q2 & -- & 34 & 28 & 49 \\
  nemotron-4-340b & NVIDIA & \checkmark & 340B & 2024-Q2 & -- & 34 & 30 & 40 \\
  gemma-2-9b & Google & \checkmark & 9B & 2024-Q2 & -- & 32 & 29 & 30 \\
  gemma-2-2b & Google & \checkmark & 2B & 2024-Q2 & -- & 30 & 27 & 24 \\
  mamba-codestral-7b-v0.1 & Mistral & \checkmark & 7B & 2024-Q3 & -- & 30 & 29 & 28 \\
  gemma-3-1b & Google & \checkmark & 1B & 2025-Q1 & -- & 26 & 26 & 28 \\
  phi-3-mini-128k & Microsoft & \checkmark & 3.8B & 2024-Q2 & -- & 25 & 25 & 24 \\
  phi-3-medium-128k & Microsoft & \checkmark & 14B & 2024-Q2 & -- & 21 & 20 & 20 \\
  \bottomrule
  \end{tabular}
  }
  \caption{Pretrained models evaluated in this study and their SimBench performance. ``Open Weights'' indicates whether model parameters are publicly released. ``Release'' reports the provider-announced release quarter. ``Reasoning'' marks models that providers explicitly market or expose as deliberate-reasoning variants. ``J-LLM-Ref-Doc'', ``J-LLM-Ref'', and ``J-LLM-Doc'' are rubric-based judge scores averaged over all task families and all three turns under three evaluation conditions (reference+documentation, reference-only, documentation-only). Models are ordered by ``J-LLM-Ref-Doc''.}
  \label{tab:model_info}
\end{table*}

\subsection{Robustness Analysis of J-LLM Metrics}

\noindent\textbf{Rubric-based J-LLM scores are robust to the choice of judge model and inference variability.}
To validate the robustness of our evaluation framework, we conducted cross-validation using three different J-LLM evaluators (GPT-4o-mini, GPT-4.1-nano, and GPT-4.1-mini) under the \textbf{J-LLM-Ref-Doc} modality, with three independent inference runs.

Table~\ref{tab:jllm_correlation} reports inter-evaluator agreement. All three evaluators exhibit high consistency, with Pearson correlations ranging from 0.8163 to 0.9489 (all $p < 0.001$). GPT-4.1-mini and GPT-4o-mini show near-perfect agreement ($r = 0.9489$), and all pairs maintain strong Spearman correlations (0.8455--0.9144), confirming consistent model rankings across evaluators.

\begin{table}[htbp]
\centering
\caption{Inter-evaluator correlation on Ref-Doc modality (all $p < 0.001$).}
\label{tab:jllm_correlation}
\begin{tabular}{lcc}
\toprule
\textbf{Evaluator Pair} & \textbf{Pearson $r$} & \textbf{Spearman $\rho$} \\
\midrule
GPT-4.1-mini vs GPT-4o-mini & 0.9489 & 0.9144 \\
GPT-4.1-mini vs GPT-4.1-nano & 0.8163 & 0.8455 \\
GPT-4.1-nano vs GPT-4o-mini & 0.8460 & 0.8905 \\
\bottomrule
\end{tabular}
\end{table}

Table~\ref{tab:jllm_statistics} summarizes the score distributions. Although the evaluators differ in absolute score levels, e.g., GPT-4.1-nano is the most lenient (mean 39.13) while GPT-4.1-mini is the most stringent (mean 32.34), the strong Pearson and Spearman correlations indicate robust cross-judge consistency. In particular, despite shifts in score calibration, the J-LLMs produce highly aligned relative rankings of generated DT solutions.

\begin{table}[htbp]
\centering
\caption{Score statistics for different J-LLMs under \textbf{J-LLM-Ref-Doc}.}
\label{tab:jllm_statistics}
\begin{tabular}{lcccc}
\toprule
\textbf{Evaluator} & \textbf{Mean} & \textbf{Std} & \textbf{Min} & \textbf{Max} \\
\midrule
GPT-4.1-mini & 32.34 & 6.59 & 14.35 & 40.62 \\
GPT-4.1-nano & 39.13 & 4.59 & 32.00 & 46.37 \\
GPT-4o-mini  & 36.00 & 5.33 & 21.75 & 40.99 \\
\bottomrule
\end{tabular}
\end{table}

Overall, these results show that the rubric-based framework yields consistent assessments across different J-LLM choices and across repeated inference runs, supporting reproducible and reliable evaluation. 

\medskip

{\noindentBold{Rubric-based J-LLM scores as practical proxies for Pass@1.}}
Execution-based correctness metrics such as Pass@1 provide the most direct validation of simulator-grade DT code, since they reflect whether the generated implementation is both syntactically valid and functionally correct in the simulator.
However, for complex multi-physics simulations, Pass@1 is costly to obtain: it requires expert assessment and/or extensive scenario-specific tests, which are difficult to author and may involve long runtimes.
A key goal of SimBench is therefore to provide an \emph{interpretable} judging signal that can serve as a practical surrogate for Pass@1 when exhaustive execution-based testing is infeasible.

We compare Pass@1 against three rubric-based J-LLM variants (Sec.~\ref{sec:modalities}): \textbf{J-LLM\_Ref\_Doc} (reference code + API excerpt), \textbf{J-LLM\_Ref} (reference code only), and \textbf{J-LLM\_Doc} (API excerpt only).
As additional baselines, we report an execution-based syntactic metric \textbf{Compile@1} and two commonly used similarity metrics, \textbf{CodeBLEU} and \textbf{ROUGE-LSUM}.
Table~\ref{tab:llm_metrics_final} summarizes the average metric values across all tasks for each S-LLM.

\begin{figure}
\centering
\includegraphics[width=0.48\textwidth]{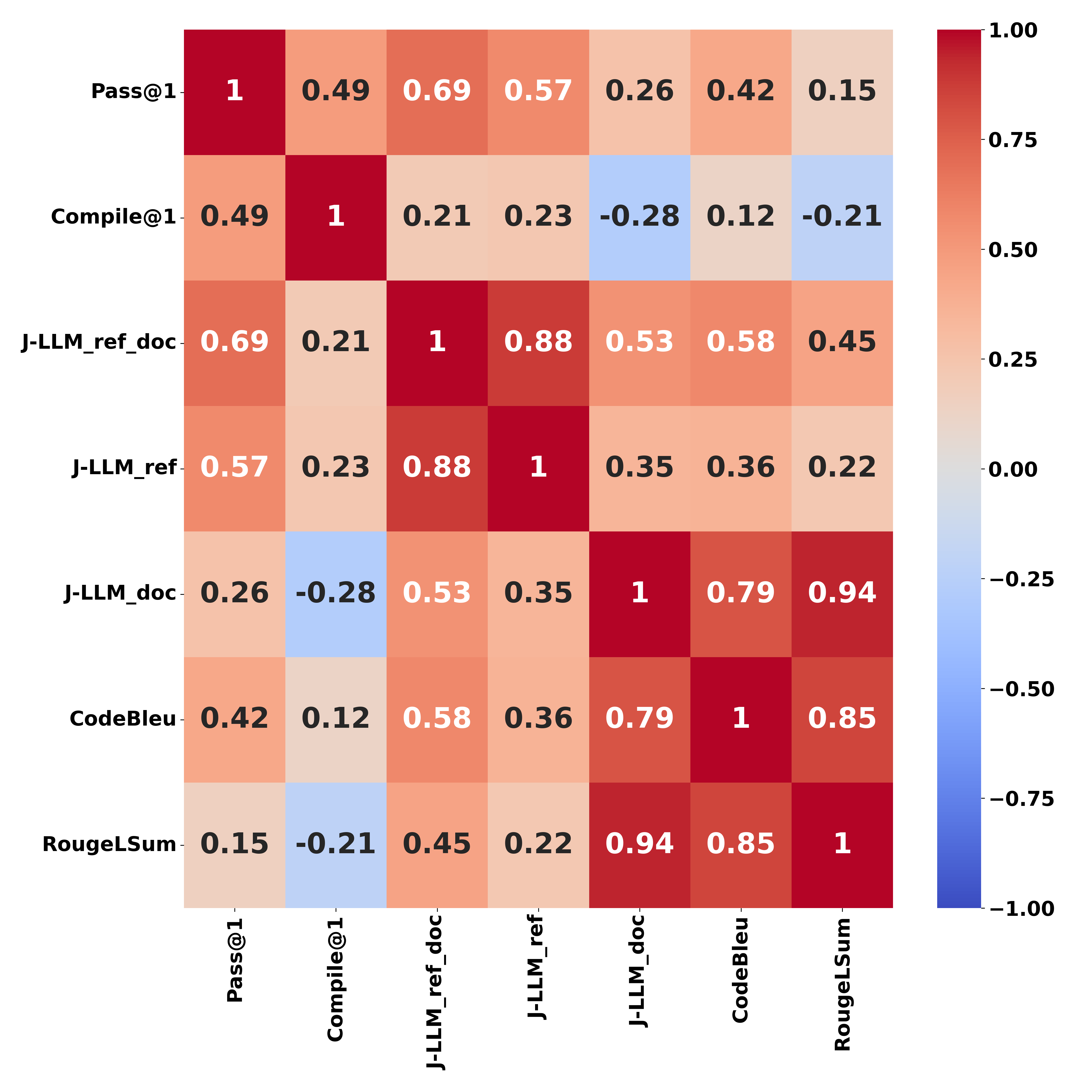}
\caption{The correlation matrix of different metrics, including pass@1, compile@1, three different J-LLM instances, and the similarity scores CodeBLEU and ROUGE-LSUM.}
\label{fig:correlation_matrix}
\end{figure}

SimBench contains $N_{\text{sys}}=34$ physical systems, each evaluated under a fixed three-turn protocol, yielding $N_{\text{task}}=102$ turn-level tasks.
We evaluate a representative set of 19 widely used models, producing
$N_{\text{inst}} = 34 \times 3 \times 19 = 1938$ matched (model, system, turn) instances with 7 metric values, total $1938 \times 7 = 13,566$ data points; see Table~\ref{tab:llm_metrics_final}.
Pass@1 is assessed by human Chrono experts, whereas Compile@1 is computed automatically.
To quantify whether rubric-based J-LLM scores track execution-based correctness, we compute Spearman rank correlations over the $N_{\text{inst}}$ matched instances across metrics.
The resulting correlation matrix is shown in Fig.~\ref{fig:correlation_matrix}.

Considering correlation with Pass@1, \textbf{J-LLM\_Ref\_Doc} achieves the strongest association ($\rho=0.69$), followed by \textbf{J-LLM\_Ref} ($\rho=0.57$) and \textbf{Compile@1} ($\rho=0.49$).
We emphasize that Compile@1 is already a strong and widely used evaluation signal in code generation, and it is often used as a reward signal in RLVR-style training.
The fact that \textbf{J-LLM\_Ref\_Doc} exhibits an even stronger correlation with Pass@1 than Compile@1 supports its effectiveness as a proxy for functional correctness.
In contrast, similarity metrics correlate more weakly with Pass@1 (CodeBLEU: $\rho=0.42$; ROUGE-LSUM: $\rho=0.15$), suggesting that surface-form overlap is insufficient to capture simulator correctness in DT generation.
We also observe that \textbf{J-LLM\_Doc} correlates strongly with similarity metrics (CodeBLEU: $\rho=0.79$; ROUGE-LSUM: $\rho=0.94$), suggesting that, under our prompting setup, a doc-only judge may behave similarly to surface-form similarity measures. Nevertheless, this can still be valuable, since similarity metrics require reference code, whereas \textbf{J-LLM\_Doc} does not.
Overall, reference-conditioned rubric judging (\textbf{J-LLM\_Ref\_Doc} and \textbf{J-LLM\_Ref}) provides a substantially stronger indicator of Pass@1 than compile- or similarity-based alternatives, supporting its use as an interpretable benchmark metric for simulator-grade DT code.

\begin{table}[htbp]
    \centering
    \resizebox{0.5\textwidth}{!}{%
    \begin{tabular}{|l|c|c|c|c|c|c|c|}
    \hline
	{\tiny \textbf{S-LLM} $\downarrow$  $|$ \textbf{Metric} $\rightarrow$} & {\tiny \textbf{Pass@1}} & {\tiny \textbf{J-LLM\_Ref\_Doc}} & {\tiny \textbf{J-LLM\_Ref}} & {\tiny \textbf{J-LLM\_Doc}} & {\tiny \textbf{ROUGE-LSUM}} & {\tiny \textbf{CodeBLEU}} & {\tiny \textbf{Compile@1}}\\
    \hline
    gpt-4o-mini & 12.8\% & 41.6 & 34.1 & 43.7 & 0.723 & 0.610 & 19.6\% \\
    codestral-22b & 9.8\% & 37.9 & 32.4 & 39.7 & 0.699 & 0.608 & 20.6\% \\
    mixtral-8x22b & 7.8\% & 38.9 & 33.8 & 36.9 & 0.667 & 0.589 & 19.6\% \\
    claude-3.5-sonnet & 6.8\% & 33.7 & 25.9 & 52.8 & 0.757 & 0.591 & 7.8\%\\
    Gemini-1.5-Pro & 6.8\% & 39.2 & 32.9 & 42.7 & 0.727 & 0.609 & 7.8\% \\
    mixtral-8x7b & 6.8\% & 37.7 & 34.4 & 30.7 & 0.624 & 0.508 & 14.7\% \\
    llama-3.1-405b & 6.8\%& 39.0 & 33.1 & 38.0 & 0.720 & 0.619 & 19.6\% \\
    mistral-nemo-12b & 6.8\% & 36.4 & 29.7 & 31.8 & 0.648 & 0.552 & 17.6\% \\
    llama-3.1-8b & 4.9\% & 37.6 & 31.1 & 34.5 & 0.656 & 0.571 & 13.7\% \\
    gemma-2-27b & 4.9\% & 37.4 & 29.3 & 36.5 & 0.710 & 0.590 & 20.6\% \\
    llama-3.1-70b & 3.9\% & 39.7 & 33.3 & 39.4 & 0.710 & 0.540 & 8.8\% \\
    gemma-2-9b & 3.9\% & 33.5 & 29.4 & 30.4 & 0.688 & 0.575 & 20.6\% \\
    mamba-codestral-7b & 2.9\% & 31.8 & 29.4 & 27.9 & 0.567 & 0.473 & 15.7\% \\
    mistral-large & 2.9\% & 37.5 & 33.3 & 43.9 & 0.740 & 0.602 & 8.8\% \\
    gemma-2-2b & 2.9\% & 31.1 & 27.5 & 24.4 & 0.641 & 0.542 & 19.6\% \\
    nemotron-4-340b & 2.9\% & 35.1 & 30.4 & 39.9 & 0.719 & 0.598 & 7.8\% \\
    phi-3-medium-128k & 2.9\% & 22.0 & 20.0 & 19.7 & 0.396 & 0.376 & 5.8\% \\
    phi-3-mini-128k & 2.9\% & 27.8 & 24.6 & 23.8 & 0.582 & 0.502 & 10.8\% \\
    gpt-4o & 2.0\% & 33.9 & 28.3 & 49.4 & 0.758 & 0.607 & 2.0\% \\
    \hline
    \end{tabular}
    }
    \caption{Comparison of S-LLMs across multiple evaluation metrics, ranked by pass@1.}
    \vspace{-10pt}
    \label{tab:llm_metrics_final}
    \end{table}
    
{\noindent\textbf{Threats to validity (LLM-as-a-judge bias).}}
While we observe strong cross-evaluator agreement, this does not rule out \emph{correlated} judge bias: LLM-based evaluators may share training-data priors that systematically favor particular coding styles, verbosity levels, or model families, yielding consistency that reflects shared bias rather than functional correctness. To mitigate this risk, we: (i) use a rubric-grounded judge prompt that prioritizes simulator-relevant, checkable criteria (e.g., API correctness, subsystem completeness, and required simulation-loop updates) and explicitly instructs the judge to ignore superficial stylistic signals (e.g., comment density and verbosity); (ii) assess ranking stability across multiple judge instances and verify alignment with human-expert rankings on a held-out subset from the calibration protocol; (iii) partially anchor rubric scores to execution-based outcomes (e.g., Pass@1) when available; and (iv) incorporate standard anti-bias prompt guardrails (e.g., position/length bias mitigation and structured scoring by rubric section). We additionally include a small adversarial-style check in which functionally correct but differently structured implementations are evaluated to test sensitivity to code organization. Nonetheless, judge scores should be interpreted as \emph{approximations} of functional correctness rather than proofs; further reducing residual bias remains important future work (e.g., increasing judge diversity, expanding executable test coverage, and adding complementary non-LLM automated checks such as static analysis and API-level validators).

\subsection{Multi-turn Analysis}

Table \ref{tab:llm_metrics_jllm} reports the multi-turn performance of S-LLMs in the SimBench environment, benchmarked with the metrics J-LLM\_Ref\_Doc. 

        \begin{table*}[htbp]
          \centering
          \resizebox{\textwidth}{!}{%
          \begin{tabular}{lcccccc}
          \toprule
          \textbf{S-LLM} & \multicolumn{5}{c}{\textbf{J-LLM Ref Doc (1st turn, 2nd turn, 3rd turn)}} & \textbf{Average} \\
          \cmidrule(lr){2-6}
          & \textbf{MBS} & \textbf{FEA} & \textbf{VEH} & \textbf{SEN} & \textbf{RBT} & \\
          \midrule
          claude-4-sonnet-20250514 & (33, 70, 57) & (37, 76, 49) & (38, 61, 51) & (34, 63, 41) & (23, 54, 47) & 49 \\
          o3 & (41, 59, 65) & (21, 70, 55) & (26, 56, 56) & (28, 51, 33) & (25, 62, 46) & 46 \\
          claude-3-7-sonnet-20250219 & (45, 58, 58) & (28, 64, 46) & (23, 56, 47) & (30, 45, 48) & (22, 52, 40) & 44 \\
          o4-mini & (34, 54, 56) & (28, 67, 44) & (26, 48, 50) & (16, 44, 42) & (27, 54, 39) & 42 \\
          qwen3-235b-a22b & (27, 51, 51) & (26, 65, 42) & (19, 50, 58) & (24, 51, 50) & (25, 56, 38) & 42 \\
          Gemini-2.5-pro & (38, 52, 46) & (30, 51, 53) & (32, 43, 48) & (28, 34, 32) & (38, 64, 28) & 41 \\
          gpt-4.1-mini & (32, 52, 50) & (27, 65, 49) & (26, 52, 47) & (22, 38, 38) & (24, 44, 48) & 41 \\
          gpt-4o-mini & (17, 67, 43) & (19, 69, 55) & (20, 52, 56) & (18, 38, 42) & (19, 59, 40) & 41 \\
          llama-3.3-70b & (21, 57, 59) & (19, 70, 46) & (20, 52, 53) & (20, 44, 37) & (17, 58, 40) & 41 \\
          llama4-maverick & (20, 63, 54) & (20, 73, 57) & (25, 50, 44) & (16, 36, 34) & (27, 58, 39) & 41 \\
          llama4-scout & (27, 58, 45) & (24, 62, 47) & (18, 50, 51) & (20, 42, 46) & (25, 61, 40) & 41 \\
          gpt-4.1-nano & (19, 55, 52) & (28, 73, 54) & (20, 45, 44) & (16, 45, 36) & (20, 55, 39) & 40 \\
          llama-3.1-70b & (21, 53, 42) & (17, 67, 50) & (19, 49, 52) & (14, 44, 47) & (33, 51, 35) & 40 \\
          Gemini-1.5-pro & (24, 55, 49) & (20, 52, 51) & (19, 52, 51) & (22, 52, 37) & (15, 50, 33) & 39 \\
          codestral-22b & (23, 56, 51) & (9, 58, 55) & (12, 43, 49) & (24, 47, 40) & (21, 58, 38) & 39 \\
          deepseek-r1-32b & (23, 67, 47) & (22, 70, 36) & (19, 48, 44) & (23, 36, 47) & (20, 50, 38) & 39 \\
          gpt-4.1 & (26, 61, 53) & (23, 62, 38) & (21, 58, 47) & (16, 47, 36) & (17, 47, 36) & 39 \\
          llama-3.1-405b & (19, 62, 54) & (20, 63, 50) & (14, 48, 51) & (15, 35, 49) & (16, 59, 36) & 39 \\
          mixtral-8x22b & (22, 63, 48) & (16, 61, 48) & (21, 41, 49) & (13, 50, 34) & (22, 60, 42) & 39 \\
          llama-3.1-8b & (10, 64, 46) & (13, 66, 49) & (18, 46, 47) & (16, 43, 36) & (16, 58, 38) & 38 \\
          mistral-nemo-12b & (17, 49, 59) & (22, 74, 51) & (15, 36, 42) & (22, 44, 44) & (25, 37, 40) & 38 \\
          mistral-large-latest & (25, 60, 43) & (10, 68, 45) & (20, 46, 44) & (26, 36, 14) & (24, 56, 40) & 37 \\
          gemma-2-27b-it & (19, 46, 46) & (25, 54, 39) & (20, 52, 52) & (19, 30, 30) & (19, 53, 31) & 36 \\
          mixtral-8x7b & (19, 62, 43) & (21, 60, 41) & (17, 58, 51) & (12, 31, 38) & (12, 44, 33) & 36 \\
          claude-3-5-sonnet & (24, 41, 34) & (25, 57, 45) & (22, 39, 39) & (27, 52, 18) & (16, 44, 27) & 34 \\
          deepseek-r1-8b & (13, 62, 45) & (13, 44, 43) & (13, 39, 48) & (13, 43, 45) & (12, 49, 35) & 34 \\
          gpt-4o & (18, 47, 50) & (18, 57, 38) & (21, 33, 43) & (22, 40, 31) & (19, 46, 33) & 34 \\
          nemotron-4-340b & (11, 50, 39) & (18, 46, 44) & (20, 48, 48) & (14, 36, 33) & (14, 50, 31) & 33 \\
          gemma-2-9b-it & (24, 43, 40) & (23, 43, 39) & (16, 42, 49) & (15, 24, 30) & (20, 53, 24) & 32 \\
          gemma-2-2b-it & (19, 41, 39) & (16, 49, 37) & (11, 40, 47) & (6, 33, 28) & (17, 38, 34) & 30 \\
          mamba-codestral-7b & (17, 47, 49) & (13, 52, 25) & (13, 48, 45) & (14, 18, 21) & (15, 45, 27) & 30 \\
          gemma-3-1b-it & (11, 25, 18) & (9, 30, 40) & (9, 47, 43) & (11, 32, 21) & (14, 42, 29) & 25 \\
          phi-3-mini-128k & (17, 27, 32) & (16, 23, 19) & (17, 41, 46) & (14, 14, 24) & (15, 46, 22) & 25 \\
          phi-3-medium-128k & (10, 43, 21) & (21, 24, 12) & (12, 24, 34) & (8, 20, 20) & (13, 37, 19) & 21 \\
          \midrule
          \textbf{Average} & (22, 53, 47) & (20, 59, 44) & (20, 47, 48) & (19, 39, 35) & (20, 51, 36) & 37 \\
          \bottomrule
          \end{tabular}
          }
          \caption{Performance of pretrained S-LLMs across five system categories (MBS, FEA, VEH, SEN, RBT) showing J-LLM Ref Doc scores for three rounds. Values in parentheses represent (1st turn, 2nd turn, 3rd turn) scores. Models are ranked by average score across all categories.}
          \vspace{-10pt}
          \label{tab:llm_metrics_jllm}
      \end{table*}
\medskip
% Section: Multi-turn Delta Analysis - Understanding Edit Capabilities
{\noindentBold{Turn-to-Turn Delta Analysis: Quantifying Edit Capabilities.}}
\label{subsec:multiturn_delta}
Iterative code modification is a critical skill in real-world development workflows. Accordingly, we analyze turn-to-turn score changes in our multi-turn evaluation to quantify how S-LLMs respond to successive modification requests. For each S-LLM -- system pair, we define:

\begin{equation}
\Delta_{12} = s_{i,2} - s_{i,1}, \quad \Delta_{23} = s_{i,3} - s_{i,2} \; ,
\end{equation}
where $s_{i,t}$ represents the J-LLM-Ref-Doc score for system $i$ at turn $t$. These deltas quantify: (1) the benefit of having existing code context ($\Delta_{12}$), and (2) the challenge of extending functionality while maintaining consistency ($\Delta_{23}$).

\textbf{Overwhelmingly positive Turn 1$\to$2 improvement.}
Table~\ref{tab:delta_global} presents the overall delta statistics across all 1,122 model-system evaluations(we have 33 S-LLMs with 34 systems). The mean $\Delta_{12} = +29.26$ is highly significant ($p < 0.001$), with \textbf{87.6\%} of cases showing improvement. This dramatic increase validates a key hypothesis: when given existing code context (Turn 2), models leverage structural information, variable naming conventions, and established patterns to produce significantly better modifications compared to generating from scratch (Turn 1). The median gain of 30 points demonstrates that this is not driven by outliers but represents a fundamental capability shift when code context is available.

\textbf{Turn 2$\to$3 reveals the difficulty of extension tasks.}
In stark contrast, $\Delta_{23} = -6.17$ is significantly negative ($p < 0.001$), with \textbf{58.5\%} of cases showing decline. This pattern reflects the increasing complexity of Turn 3 tasks, which require: (1) extending existing functionality (adding new features), (2) maintaining consistency with previous modifications, and (3) preserving correctness of unchanged components.

\begin{table}[htbp]
    \centering
    \caption{Global turn-to-turn delta statistics across 1,122 model-system pairs. All differences are highly significant ($p < 0.001$).}
    \label{tab:delta_global}
    \begin{tabular}{lcccc}
    \toprule
    \textbf{Delta} & \textbf{Mean} & \textbf{Median} & \textbf{IQR} & \textbf{Improvement \%} \\
    \midrule
    $\Delta_{12}$ & $+29.26 $ & $+30.0$ & $[13.3, 46.0]$ & 87.6\% \\
    $\Delta_{23}$ & $-6.17 $ & $-7.0$ & $[-25.0, 11.0]$ & 38.2\% \\
    \bottomrule
    \end{tabular}
\end{table}

\textbf{Implications for real-world deployment.}
These findings have critical implications for deploying S-LLMs in interactive development environments:

\begin{itemize}
    \item \textbf{Leverage existing code:} The substantial $\Delta_{12}$ improvements (+30 points, 87.6\% success rate) strongly advocate for always providing existing code context when requesting modifications. Tools should default to showing relevant surrounding code rather than asking models to generate modifications in isolation.
    
    \item \textbf{Manage complexity in extensions:} The consistent $\Delta_{23}$ decline (-6.2 points, 58.5\% degradation rate) suggests that large feature additions should be broken into smaller incremental steps. Rather than requesting complex multi-part extensions in a single turn, better results may come from iterative refinement with validation checkpoints.
\end{itemize}

This delta analysis indicates that SimBench probes multi-turn code editing and extension behavior, not just one-shot code generation. The resulting turn-to-turn trends—context sensitivity, extension difficulty, and category-dependent behavior, provide a diagnostic view of current S-LLM strengths and failure modes.

%\begin{figure}
%\centering
%\includegraphics[width=0.4\textwidth]{images/pass_compile.png}
%\caption{The performance of S-LLMs in the SimBench environment, benchmarked with the metrics pass@1 and compile@1.}
%\end{figure}

%\begin{figure}
%\centering
%includegraphics[width=0.4\textwidth]{images/llm_scores.png}
%\caption{The performance of various S-LLMs in the SimBench environment, benchmarked with J-LLMs using three different strategies.}
%\end{figure}

\subsection{Failure Mode Analysis: System Difficulty Assessment}
\label{subsec:failure_mode}

To identify challenging scenarios and failure patterns in simulation code generation, we conduct a comprehensive system-level difficulty analysis. Table~\ref{tab:system_difficulty} presents the performance of all 34 simulation systems, ranked by overall difficulty (average J-LLM-Ref-Doc scores across all S-LLMs and all three turns). Lower scores indicate harder systems where models consistently struggle. We highlight next the lessons learned from this analysis.

\textbf{Sensor systems emerge as the most challenging domain.}
The hardest system overall is \texttt{lidar} (SEN category, overall score 23.4), which requires intricate sensor configuration including noise models, visualization pipelines, and dynamic sensor positioning. The SEN category achieves the lowest average performance (31.3), significantly below other domains. This persistent difficulty stems from:

\begin{itemize}
    \item \textbf{API complexity}: Sensor initialization requires precise parameter tuning (e.g., resolution, field-of-view, update rates) with limited error tolerance.
    \item \textbf{Visualization challenges}: Sensors demand additional rendering setup, buffer management, and frame processing that are domain-specific and poorly documented in training data.
    \item \textbf{Real-time constraints}: Sensor simulations require careful synchronization between physics updates and sensor data acquisition, a subtle requirement often missed by models.
\end{itemize}

Similarly, \texttt{camera} (SEN, 30.1) exhibits consistently low performance across all turns (T1=18.3, T2=45.8, T3=26.0), with the Turn 2$\to$3 decline (-19.7)
 highlighting the difficulty of extending sensor functionality while maintaining correct configuration. An illustrative prompt exemplifies these challenges:

\begin{quote}
\textit{Create a PyChrono simulation where a triangular mesh (loaded from a Wavefront .obj file) is visualized as a fixed body in the scene. Add a camera sensor to the body, managed by a sensor manager, with noise filters and visualizations applied to the camera images. Simulate the system, dynamically updating the camera's position in an orbit around the mesh, and print out camera buffer data at each step.}
\end{quote}

This complexity significantly contributes to lower performance metrics observed in sensor-based scenarios.

\medskip

\textbf{Robotics and vehicle systems show moderate difficulty with high variability.}
The RBT category (average 35.8) and VEH category (38.0) occupy the middle difficulty range. Systems like \texttt{vehros} (RBT, 27.8) and \texttt{sedan} (VEH, 28.6) demonstrate interesting failure patterns: relatively low Turn 1 generation scores but substantial Turn 2 improvements (+19.1 and +16.2 respectively), suggesting models can leverage context effectively once initial code structure is provided. However, the high standard deviation in these categories (e.g., \texttt{art} with std=32.1) indicates significant performance variance across different models—some handle complex kinematics well while others struggle with constraint specification.

\textbf{FEA and MBS systems are relatively easier but exhibit distinct patterns.}
The FEA category achieves the highest average performance (41.0), with \texttt{art} reaching 58.6, the easiest system overall. This success likely reflects: (1) well-structured simulation workflows with clear mesh$\to$material$\to$solver pipelines, (2) abundant finite element examples in training corpora, and (3) less ambiguity in problem specifications. The MBS category (40.8) also performs well, though systems like \texttt{cable} exhibit dramatic Turn 3 declines (-37.8), revealing challenges in extending multi-body dynamics while preserving kinematic consistency.

\textbf{Turn-specific failure modes reveal distinct challenges.}
Turn 1 (generation from scratch) proves universally difficult, with scores averaging only 20.2 and systems like \texttt{vehros} scoring just 14.7. This baseline difficulty underscores the challenge of synthesizing complete simulation setups without examples. Turn 2 (modification with context) shows dramatic improvement (average +29.3), but the benefit varies widely—from +8.7 for systems with template-like structures (minimal leverage from context) to +64.3 for complex systems where context provides critical scaffolding. Turn 3 (extension tasks) reveals brittleness: 22 out of 34 systems show performance declines, with some catastrophic drops (e.g., \texttt{cable}: -37.8) indicating models struggle to add functionality while maintaining simulation correctness.

\textbf{Implications for S-LLM development.}
While S-LLMs handle well simple, incremental edits, they still struggle with complex simulation DT generation, particularly for sensor-centric tasks. SimBench makes these limitations visible in a controlled, multi-turn setting. Ultimately, our failure mode analysis identifies three priority areas for improvement:

\begin{enumerate}
    \item \textbf{Sensor domain specialization}: Targeted training on sensor configuration patterns, visualization pipelines, and buffer management could significantly improve the weakest category.
    \item \textbf{Consistency-preserving extensions}: The widespread Turn 3 degradation calls for methods that explicitly verify backward compatibility when extending code (e.g., constraint checking, regression testing).
    \item \textbf{Variance reduction}: High inter-model variance in complex systems (robotics, vehicles) suggests ensemble or hybrid approaches might provide more reliable generation.
\end{enumerate}

These findings complement our multi-turn delta analysis (Section~\ref{subsec:multiturn_delta}), providing system-level granularity that guides both model selection for specific simulation domains and targeted improvements in S-LLM capabilities.

\begin{table}[htbp]
    \centering
    \caption{System difficulty analysis: average J-LLM-Ref-Doc scores across all S-LLMs and turns. Systems ranked by overall difficulty (lower scores indicate harder systems).}
    \label{tab:system_difficulty}
    \resizebox{0.5\textwidth}{!}{%
    \begin{tabular}{rlccccc}
    \toprule
    \textbf{Rank} & \textbf{System} & \textbf{Category} & \textbf{Turn 1} & \textbf{Turn 2} & \textbf{Turn 3} & \textbf{Overall} \\
    \midrule
    1 & lidar & SEN & 19.5 & 31.3 & 19.3 & 23.4 \\
    2 & vehros & RBT & 14.7 & 33.9 & 34.7 & 27.8 \\
    3 & sedan & VEH & 21.9 & 38.1 & 25.9 & 28.6 \\
    4 & camera & SEN & 18.3 & 45.8 & 26.0 & 30.1 \\
    5 & scm\_hill & VEH & 19.9 & 28.6 & 43.1 & 30.5 \\
    6 & veh\_app & SEN & 19.9 & 36.5 & 35.4 & 30.6 \\
    7 & m113 & VEH & 17.8 & 36.0 & 39.0 & 30.9 \\
    8 & kraz & VEH & 18.8 & 54.8 & 21.7 & 31.8 \\
    9 & scm & VEH & 22.5 & 43.7 & 32.7 & 32.9 \\
    10 & feda & VEH & 16.5 & 43.4 & 39.5 & 33.1 \\
    11 & sensros & RBT & 17.5 & 32.2 & 50.2 & 33.3 \\
    12 & pendulum & MBS & 19.7 & 39.7 & 41.0 & 33.5 \\
    13 & buckling & FEA & 21.0 & 53.2 & 28.4 & 34.2 \\
    14 & hmmwv & VEH & 21.9 & 31.5 & 50.9 & 34.8 \\
    15 & uazbus & VEH & 15.3 & 57.6 & 33.5 & 35.5 \\
    16 & handler & RBT & 20.1 & 59.5 & 27.6 & 35.7 \\
    17 & man & VEH & 20.1 & 53.1 & 34.6 & 35.9 \\
    18 & turtlebot & RBT & 20.8 & 61.6 & 29.2 & 37.2 \\
    19 & gear & MBS & 21.4 & 57.3 & 39.0 & 39.3 \\
    20 & viper & RBT & 22.6 & 66.4 & 30.2 & 39.7 \\
    21 & rigid\_multipatches & VEH & 20.7 & 35.9 & 63.7 & 40.1 \\
    22 & tablecloth & FEA & 22.3 & 51.7 & 47.6 & 40.5 \\
    23 & curiosity & RBT & 26.3 & 55.0 & 41.3 & 40.9 \\
    24 & gps\_imu & SEN & 18.6 & 44.3 & 60.5 & 41.1 \\
    25 & gator & VEH & 21.0 & 36.8 & 68.1 & 42.0 \\
    26 & cable & FEA & 21.1 & 71.7 & 33.9 & 42.3 \\
    27 & mass\_spring\_damper & MBS & 23.9 & 56.2 & 47.1 & 42.4 \\
    28 & slider\_crank & MBS & 27.2 & 52.4 & 50.2 & 43.3 \\
    29 & rotor & FEA & 19.8 & 62.8 & 48.3 & 43.6 \\
    30 & beam & FEA & 17.8 & 53.6 & 61.5 & 44.3 \\
    31 & particles & MBS & 19.6 & 61.4 & 55.3 & 45.4 \\
    32 & citybus & VEH & 19.6 & 50.4 & 67.4 & 45.8 \\
    33 & rigid\_highway & VEH & 21.2 & 63.6 & 70.0 & 51.6 \\
    34 & art & VEH & 17.9 & 82.2 & 75.8 & 58.6 \\
    \bottomrule
    \end{tabular}
    }
\end{table}

\section{Conclusion and Future Work}
\label{sec:conclusion}

%we should grow the dataset used to improve our J-LLM，including more dynamics sytems e.g. fluid-solid interaction; multi-phrase flow; 

This work introduces SimBench, a benchmark and dataset designed to evaluate an S-LLM's ability to generate multi-turn digital twins (DTs) for multi-physics simulation. Despite broad exposure of modern S-LLMs to public simulation code, the strongest out-of-the-box model in our study attains only Pass@1 = 13\% on SimBench, highlighting substantial headroom and positioning SimBench to guide future improvements for Chrono DT generation as well as for other simulation ecosystems. We further show that the proposed rubric-based J-LLM provides a more informative and effective evaluation signal than execution-only proxies such as Compile@1 and text-similarity metrics (e.g., CodeBLEU and ROUGE-LSUM), as it better tracks Pass@1, our operational measure of end-to-end functional correctness. In addition, incorporating reference implementations and API documentation improves J-LLM judging performance, suggesting promising directions for future work, including retrieval-augmented judging and targeted fine-tuning using expert references and documentation. Beyond expanding SimBench to additional scenario classes (e.g., fluid--solid interaction and multi-phase flow), a key next step is to port the benchmark to other simulators by re-instantiating the task suite, documentation/reference context, and execution oracle.

\FloatBarrier

\subsection*{Reproducibility \& Artifacts.}
To support independent verification and enable standardized comparisons, SimBench is released as an open artifact package at \texttt{https://github.com/uwsbel/SimBench}. The release includes the benchmark specification (task definitions and metadata), the prompts/rubrics used for multi-turn evaluation, and the complete evaluation pipeline (scripts to run scoring and aggregate turn- and system-level results), together with an environment specification to reproduce the software stack. Beyond a pointer to the benchmark, these artifacts make the experimental protocol explicit and auditable, reduce re-implementation variance across groups, and provide a concrete substrate for ablations (e.g., prompting or rubric changes) and for extending SimBench to additional systems while preserving comparability.

\bibliographystyle{IEEEtran}
\bibliography{BibFiles/refsML-AI,BibFiles/refsMBS,BibFiles/refsChronoSpecific,BibFiles/refsCompSci,BibFiles/refsOddsEnds,BibFiles/refsTerramech,BibFiles/refsFSI,BibFiles/refsRobotics}

\begin{appendices}
    \section{Prompt Templates Used to Query S-LLMs}
\label{app:prompt_templates}

\noindent The templates below are fixed across all evaluated S-LLMs to ensure reproducibility; only the bracketed fields (\texttt{\{first\_prompt\}}, \texttt{\{code\}}, \texttt{\{prompt\}}) are instantiated per task/turn.

\begin{PromptBlock}{Turn 1 Wrapper Prompt (DT construction from a vague request)}
You are a PyChrono expert tasked with generating a simulation script based on the following instructions. Make sure to:
(*@\textbf{1.}@*) Initialize the PyChrono environment and core components.\\
(*@\textbf{2.}@*) Add the required physical systems and objects as specified.\\
(*@\textbf{3.}@*) Set necessary default parameters such as positions, forces, and interactions when not explicitly provided.\\
(*@\textbf{4.}@*) Include a simulation loop (and visualization if requested).\\
(*@\textbf{5.}@*) Output only the final runnable Python script (no explanations, no markdown).

Instructions:
(*@\texttt{\{first\_prompt\}}@*)
\end{PromptBlock}

\begin{PromptBlock}{Turns 2--3 Wrapper Prompt (code editing with bug fixing and sharp modifications)}
You are a PyChrono expert tasked with generating a simulation script based on the following instructions and a given PyChrono script, which may contain errors.

Your task has two parts:
(*@\textbf{(i)}@*) Identify and correct any issues that would prevent correct execution (including syntax errors, logical errors, incorrect method names, and parameter issues).\\
(*@\textbf{(ii)}@*) Modify the script to meet the new requirements.

Preserve existing functionality unless it directly conflicts with the new requirements.

IMPORTANT: Output only the final corrected and modified runnable Python script (no explanations, no markdown).

Here is the PyChrono code you need to modify:
(*@\texttt{\{code\}}@*)

Please modify the given code based on the following instructions:
(*@\texttt{\{prompt\}}@*)
\end{PromptBlock}

\section{J-LLM Prompt Templates}
\label{app:jllm_prompts}

\noindent The following templates are used to query the rubric-based judge model (J-LLM). 
Bracketed fields (\texttt{\{code\}}, \texttt{\{reference\_code\}}, \texttt{\{api\_documentation\_link\}}) are instantiated per task/turn.

% ---------------------------
% J-LLM_Ref_Doc
% ---------------------------
\begin{PromptBlock}{J-LLM\_Ref\_Doc (Reference + API Documentation)}
You are a PyChrono expert tasked with evaluating a simulation script by comparing it against a reference script generated by experts. Your evaluation should consider both the accuracy of the script compared to the reference and adherence to best practices as outlined in the PyChrono API documentation.

Here is the PyChrono code you need to evaluate:
[The Start of Assistant's Answer]
(*@\texttt{\{code\}}@*)
[The End of Assistant's Answer]

Here is the expert-generated reference code:
[The Start of Reference Answer]
(*@\texttt{\{reference\_code\}}@*)
[The End of Reference Answer]

Use the following evaluation criteria and point deduction guidelines:

(*@\textbf{1. Completeness (40 points total)}@*)
- Compare the provided code to the reference script. Deduct (*@\textbf{15 points}@*) for each missing essential component (e.g., system initialization, body creation, visualization) that is present in the reference script.
- Deduct (*@\textbf{10 points}@*) if a component is present but lacks important details or is incorrectly configured compared to the reference.
- Deduct (*@\textbf{5 points}@*) for minor omissions or slight deviations from the reference.

(*@\textbf{2. Correctness (30 points total)}@*)
- Compare the code to the reference script. Deduct (*@\textbf{15 points}@*) for each incorrect use of a PyChrono API that could lead to a significant change in simulation behavior.
- Deduct (*@\textbf{10 points}@*) for logical errors in the code, such as incorrect joint initialization or wrong setting of body properties, especially if the reference script does it correctly.
- Deduct (*@\textbf{5 points}@*) for minor inaccuracies or unnecessary API calls that deviate from the reference script.

(*@\textbf{3. Code Quality (10 points total)}@*)
- Evaluate the readability, structure, and documentation of the code against the reference script. Deduct (*@\textbf{5--10 points}@*) for poor readability, structure, or lack of meaningful variable names and formatting.
- Deduct (*@\textbf{5 points}@*) for insufficient comments or failure to follow documentation best practices, especially if the reference script provides better documentation.

(*@\textbf{4. Efficiency (10 points total)}@*)
- Evaluate the efficiency of the code compared to the reference script. Deduct (*@\textbf{5 points}@*) for each instance of unnecessary calculations, redundant code, or inefficient use of APIs that is optimized in the reference script.
- Deduct (*@\textbf{3 points}@*) for missing obvious optimization opportunities that the reference script implements.

(*@\textbf{5. Error Handling and Robustness (5 points total)}@*)
- Assess the error handling and robustness of the code. Deduct (*@\textbf{5 points}@*) for lack of basic error handling or failure to account for common issues that the reference script handles.
- Deduct (*@\textbf{3 points}@*) for inadequate handling of edge cases compared to the reference script.

(*@\textbf{6. Use of Visualization Tools (5 points total)}@*)
- Compare the use of visualization tools in the provided code to the reference script. Deduct (*@\textbf{3--5 points}@*) for incorrect or inadequate visualization setup as per the reference script.
- Deduct (*@\textbf{2 points}@*) for minor visualization issues, such as suboptimal lighting or incomplete setup of visual elements, compared to the reference.

Avoid position biases and ensure that the order in which the responses are presented does not influence your decision. Do not allow the length of the responses to influence your evaluation. Do not favor certain names of the assistants. Be as objective as possible.

Reference the PyChrono API documentation provided here: (*@\texttt{\{api\_documentation\_link\}}@*)

After providing your explanation, output the final score using the following format: (*@\textbf{[[x]]}@*) where x is the score assigned to the assistant's answer.

Provide the evaluated score and a brief explanation of the deductions below:
\end{PromptBlock}

% ---------------------------
% J-LLM_Ref
% ---------------------------
\begin{PromptBlock}{J-LLM\_Ref (Reference Only)}
You are a PyChrono expert tasked with evaluating a simulation script by comparing it against a reference script generated by experts.

Here is the PyChrono code you need to evaluate:
[The Start of Assistant's Answer]
(*@\texttt{\{code\}}@*)
[The End of Assistant's Answer]

Here is the expert-generated reference code:
[The Start of Reference Answer]
(*@\texttt{\{reference\_code\}}@*)
[The End of Reference Answer]

Use the following evaluation criteria and point deduction guidelines:

(*@\textbf{1. Completeness (40 points total)}@*)
- Compare the provided code to the reference script. Deduct (*@\textbf{15 points}@*) for each missing essential component (e.g., system initialization, body creation, visualization) that is present in the reference script.
- Deduct (*@\textbf{10 points}@*) if a component is present but lacks important details or is incorrectly configured compared to the reference.
- Deduct (*@\textbf{5 points}@*) for minor omissions or slight deviations from the reference script.

(*@\textbf{2. Correctness (30 points total)}@*)
- Compare the code to the reference script. Deduct (*@\textbf{15 points}@*) for each incorrect use of a PyChrono API that could lead to a significant change in simulation behavior.
- Deduct (*@\textbf{10 points}@*) for logical errors in the code, such as incorrect joint initialization or wrong setting of body properties, especially if the reference script does it correctly.
- Deduct (*@\textbf{5 points}@*) for minor inaccuracies or unnecessary API calls that deviate from the reference script.

(*@\textbf{3. Code Quality (10 points total)}@*)
- Evaluate the readability, structure, and documentation of the code against the reference script. Deduct (*@\textbf{5--10 points}@*) for poor readability, structure, or lack of meaningful variable names and formatting.
- Deduct (*@\textbf{5 points}@*) for insufficient comments or failure to follow documentation best practices, especially if the reference script provides better documentation.

(*@\textbf{4. Efficiency (10 points total)}@*)
- Evaluate the efficiency of the code compared to the reference script. Deduct (*@\textbf{5 points}@*) for each instance of unnecessary calculations, redundant code, or inefficient use of APIs that is optimized in the reference script.
- Deduct (*@\textbf{3 points}@*) for missing obvious optimization opportunities that the reference script implements.

(*@\textbf{5. Error Handling and Robustness (5 points total)}@*)
- Assess the error handling and robustness of the code. Deduct (*@\textbf{5 points}@*) for lack of basic error handling or failure to account for common issues that the reference script handles.
- Deduct (*@\textbf{3 points}@*) for inadequate handling of edge cases compared to the reference script.

(*@\textbf{6. Use of Visualization Tools (5 points total)}@*)
- Compare the use of visualization tools in the provided code to the reference script. Deduct (*@\textbf{3--5 points}@*) for incorrect or inadequate visualization setup as per the reference script.
- Deduct (*@\textbf{2 points}@*) for minor visualization issues, such as suboptimal lighting or incomplete setup of visual elements, compared to the reference.

Avoid position biases and ensure that the order in which the responses are presented does not influence your decision. Do not allow the length of the responses to influence your evaluation. Do not favor certain names of the assistants. Be as objective as possible.

After providing your explanation, output the final score using the following format: (*@\textbf{[[x]]}@*) where x is the score assigned to the assistant's answer.

Provide the evaluated score and a brief explanation of the deductions below:
\end{PromptBlock}

% ---------------------------
% J-LLM_Doc
% ---------------------------
\begin{PromptBlock}{J-LLM\_Doc (API Documentation Only)}
You are a PyChrono expert tasked with evaluating a simulation script by comparing it against the PyChrono API documentation. While the API documentation provides guidelines, it may not cover all aspects due to length constraints. Therefore, your evaluation should also be based on your knowledge of best practices in Python coding and general simulation principles.

Here is the PyChrono code you need to evaluate:
[The Start of Assistant's Answer]
(*@\texttt{\{code\}}@*)
[The End of Assistant's Answer]

Use the following evaluation criteria and point deduction guidelines:

(*@\textbf{1. Completeness (40 points total)}@*)
- Deduct (*@\textbf{15 points}@*) for each missing essential component (e.g., system initialization, body creation, visualization) as outlined in the PyChrono API documentation or generally expected in a simulation setup.
- Deduct (*@\textbf{10 points}@*) if a component is present but lacks important details or is incorrectly configured according to the API documentation or general simulation best practices.
- Deduct (*@\textbf{5 points}@*) for minor omissions or slight deviations from best practices mentioned in the API documentation or common Python coding practices.

(*@\textbf{2. Correctness (30 points total)}@*)
- Deduct (*@\textbf{15 points}@*) for each incorrect use of a PyChrono API that could lead to a significant change in simulation behavior, as indicated by the documentation or your expert knowledge.
- Deduct (*@\textbf{10 points}@*) for logical errors in the code, such as incorrect joint initialization or wrong setting of body properties, based on the API documentation or standard simulation principles.
- Deduct (*@\textbf{5 points}@*) for minor inaccuracies or unnecessary API calls that deviate from the API guidelines or standard coding practices.

(*@\textbf{3. Code Quality (10 points total)}@*)
- Evaluate the readability, structure, and documentation of the code. Deduct (*@\textbf{5--10 points}@*) for poor readability, structure, or lack of meaningful variable names and formatting, based on your Python expertise.
- Deduct (*@\textbf{5 points}@*) for insufficient comments or failure to follow documentation best practices, whether outlined in the API documentation or based on general coding standards.

(*@\textbf{4. Efficiency (10 points total)}@*)
- Deduct (*@\textbf{5 points}@*) for each instance of unnecessary calculations, redundant code, or inefficient use of APIs that could be optimized according to the API documentation or your understanding of efficient coding practices.
- Deduct (*@\textbf{3 points}@*) for missing obvious optimization opportunities as suggested by the API documentation or standard programming practices.

(*@\textbf{5. Error Handling and Robustness (5 points total)}@*)
- Deduct (*@\textbf{5 points}@*) for lack of basic error handling or failure to account for common issues, as recommended by the API documentation or best practices in Python coding.
- Deduct (*@\textbf{3 points}@*) for inadequate handling of edge cases, considering both the API documentation and typical robustness requirements in coding.

(*@\textbf{6. Use of Visualization Tools (5 points total)}@*)
- Deduct (*@\textbf{3--5 points}@*) for incorrect or inadequate visualization setup according to the API documentation or general expectations for visualizing simulations.
- Deduct (*@\textbf{2 points}@*) for minor visualization issues, such as suboptimal lighting or incomplete setup of visual elements, based on both the API documentation and your understanding of effective simulation visualization.

Avoid position biases and ensure that the order in which the responses are presented does not influence your decision. Do not allow the length of the responses to influence your evaluation. Do not favor certain names of the assistants. Be as objective as possible.

Reference the PyChrono API documentation provided here: (*@\texttt{\{api\_documentation\_link\}}@*)

After providing your explanation, output the final score using the following format: (*@\textbf{[[x]]}@*) where x is the score assigned to the assistant's answer.

Provide the evaluated score and a brief explanation of the deductions below:
\end{PromptBlock}

\section{Curiosity Rover Example: Ground-Truth (Expert Reference) Code}
\label{app:curiosity_gt}

\noindent The following scripts are expert-authored reference implementations for the Curiosity rover example used in paragraph~\ref{app:curiosity_example}. They correspond to Turns~1--3 respectively.

\begin{CodeBlock}{Curiosity (RBT) --- Turn 1 Ground Truth (Expert Reference)}
import os
import math
import numpy as np
import pychrono as chrono
import pychrono.robot as robot
from pychrono import irrlicht as chronoirr

# Create Chrono system
system = chrono.ChSystemNSC()
system.SetCollisionSystemType(
    chrono.ChCollisionSystem.Type_BULLET)
system.SetGravitationalAcceleration(chrono.ChVector3d(0, 0, -9.81))
chrono.ChCollisionModel.SetDefaultSuggestedEnvelope(
    0.0025)
chrono.ChCollisionModel.SetDefaultSuggestedMargin(0.0025)

# Create ground body with contact material and add it to the system
ground_mat = chrono.ChContactMaterialNSC()
ground = chrono.ChBodyEasyBox(20, 20, 1, 1000, True, True, ground_mat)
ground.SetPos(chrono.ChVector3d(0, 0, -0.5))  # Position the ground slightly below the origin
ground.SetFixed(True)  # Fix the ground in place
ground.GetVisualShape(0).SetTexture(
    chrono.GetChronoDataFile("textures/concrete.jpg"))
system.Add(ground)

# Create Curiosity rover and add it to the system
rover = robot.Curiosity(system)

# Create driver for rover
driver = robot.CuriosityDCMotorControl()
rover.SetDriver(driver)

# Initialize rover position and orientation
init_pos = chrono.ChVector3d(0, 0.2, 0)
init_rot = chrono.ChQuaterniond(1, 0, 0, 0)
rover.Initialize(chrono.ChFramed(init_pos, init_rot))

# Create the Irrlicht visualization
vis = chronoirr.ChVisualSystemIrrlicht()
vis.AttachSystem(system)
vis.SetCameraVertical(chrono.CameraVerticalDir_Z)
vis.SetWindowSize(1280, 720)
vis.SetWindowTitle("Curiosity rover - Rigid terrain")
vis.Initialize()
vis.AddLogo(chrono.GetChronoDataFile(
    "logo_pychrono_alpha.png"))
vis.AddSkyBox()
vis.AddCamera(chrono.ChVector3d(0, 3, 3), chrono.ChVector3d(0, 0, 0))
vis.AddTypicalLights()
vis.AddLightWithShadow(chrono.ChVector3d(1.5, -2.5, 5.5), chrono.ChVector3d(0, 0, 0), 3, 4, 10, 40, 512)

# Enable shadows (commented out to improve performance)
# vis.EnableShadows()

# Set the simulation time step
time_step = 1e-3

# Simulation loop
time = 0
while vis.Run():
    time += time_step

    # Set steering input for the rover after 1 second
    steering = 0
    if time >= 1:
        steering = (time - 1) * 0.2
    driver.SetSteering(steering)

    # Update rover dynamics
    rover.Update()

    # Render the scene
    vis.BeginScene()
    vis.Render()
    vis.EndScene()

    # Advance simulation by one time step
    system.DoStepDynamics(time_step)
\end{CodeBlock}

\begin{CodeBlock}{Curiosity (RBT) --- Turn 2 Ground Truth (Expert Reference)}
import os
import math
import numpy as np
import pychrono as chrono
import pychrono.robot as robot
from pychrono import irrlicht as chronoirr

# Create Chrono system
system = chrono.ChSystemNSC()
system.SetCollisionSystemType(
    chrono.ChCollisionSystem.Type_BULLET)
system.SetGravitationalAcceleration(chrono.ChVector3d(0, 0, -9.81))
chrono.ChCollisionModel.SetDefaultSuggestedEnvelope(
    0.0025)
chrono.ChCollisionModel.SetDefaultSuggestedMargin(0.0025)

# Create ground body with contact material and add it to the system
ground_mat = chrono.ChContactMaterialNSC()
ground = chrono.ChBodyEasyBox(20, 20, 1, 1000, True, True, ground_mat)
ground.SetPos(chrono.ChVector3d(0, 0, -0.5))  # Position the ground slightly below the origin
ground.SetFixed(True)  # Fix the ground in place
ground.GetVisualShape(0).SetTexture(
    chrono.GetChronoDataFile("textures/concrete.jpg"))
system.Add(ground)

# create a long box for rover to cross
box = chrono.ChBodyEasyBox(0.25, 5, 0.25, 1000, True, True, ground_mat)
box.SetPos(chrono.ChVector3d(0, 0, 0.0))
box.SetFixed(True)
box.GetVisualShape(0).SetTexture(
    chrono.GetChronoDataFile("textures/blue.png"))
system.Add(box)

# Create Curiosity rover and add it to the system
rover = robot.Curiosity(system)

# Create driver for rover
driver = robot.CuriosityDCMotorControl()
rover.SetDriver(driver)

# Initialize rover position and orientation
init_pos = chrono.ChVector3d(-5, 0.0, 0)
init_rot = chrono.ChQuaterniond(1, 0, 0, 0)
rover.Initialize(chrono.ChFramed(init_pos, init_rot))

# Create the Irrlicht visualization
vis = chronoirr.ChVisualSystemIrrlicht()
vis.AttachSystem(system)
vis.SetCameraVertical(chrono.CameraVerticalDir_Z)
vis.SetWindowSize(1280, 720)
vis.SetWindowTitle("Curiosity rover - Rigid terrain")
vis.Initialize()
vis.AddLogo(chrono.GetChronoDataFile(
    "logo_pychrono_alpha.png"))
vis.AddSkyBox()
vis.AddCamera(chrono.ChVector3d(0, 3, 3), chrono.ChVector3d(0, 0, 0))
vis.AddTypicalLights()
vis.AddLightWithShadow(chrono.ChVector3d(1.5, -2.5, 5.5), chrono.ChVector3d(0, 0, 0), 3, 4, 10, 40, 512)

# Enable shadows (commented out to improve performance)
# vis.EnableShadows()
# Set the simulation time step
time_step = 1e-3
# Simulation loop
time = 0
while vis.Run():
    time += time_step
    # ask rover to move forward
    driver.SetSteering(0.0)

    # Update rover dynamics
    rover.Update()

    # Render the scene
    vis.BeginScene()
    vis.Render()
    vis.EndScene()

    # Advance simulation by one time step
    system.DoStepDynamics(time_step)
\end{CodeBlock}

\begin{CodeBlock}[breakable]{Curiosity (RBT) --- Turn 3 Ground Truth (Expert Reference)}
import os
import math
import numpy as np
import pychrono as chrono
import pychrono.robot as robot
import pychrono.sensor as sens
from pychrono import irrlicht as chronoirr

# Create Chrono system
system = chrono.ChSystemNSC()
system.SetCollisionSystemType(
    chrono.ChCollisionSystem.Type_BULLET)
system.SetGravitationalAcceleration(chrono.ChVector3d(0, 0, -9.81))
chrono.ChCollisionModel.SetDefaultSuggestedEnvelope(
    0.0025)
chrono.ChCollisionModel.SetDefaultSuggestedMargin(0.0025)

# Create ground body with contact material and add it to the system
ground_mat = chrono.ChContactMaterialNSC()
ground = chrono.ChBodyEasyBox(20, 20, 1, 1000, True, True, ground_mat)
ground.SetPos(chrono.ChVector3d(0, 0, -0.5))  # Position the ground slightly below the origin
ground.SetFixed(True)  # Fix the ground in place
ground.GetVisualShape(0).SetTexture(
    chrono.GetChronoDataFile("textures/concrete.jpg"))
system.Add(ground)

# create a long box for rover to cross
box = chrono.ChBodyEasyBox(0.25, 5, 0.25, 1000, True, True, ground_mat)
box.SetPos(chrono.ChVector3d(0, 0, 0.0))
box.SetFixed(True)
box.GetVisualShape(0).SetTexture(
    chrono.GetChronoDataFile("textures/blue.png"))
system.Add(box)

# Create Curiosity rover and add it to the system
rover = robot.Curiosity(system)

# Create driver for rover
driver = robot.CuriosityDCMotorControl()
rover.SetDriver(driver)

# Initialize rover position and orientation
init_pos = chrono.ChVector3d(-5, 0.0, 0)
init_rot = chrono.ChQuaterniond(1, 0, 0, 0)
rover.Initialize(chrono.ChFramed(init_pos, init_rot))

# Create sensor manager
manager = sens.ChSensorManager(system)

# Create the lidar sensor and attach it to the rover
offset_pose = chrono.ChFramed(
        chrono.ChVector3d(3.0, 0, 1), chrono.QuatFromAngleAxis(0, chrono.ChVector3d(0, 1, 0))
    )

# lidar related parameters
# Update rate in Hz
update_rate = 5.0
# Number of horizontal and vertical samples
horizontal_samples = 800
vertical_samples = 300
# Horizontal and vertical field of view (radians)
horizontal_fov = 2 * chrono.CH_PI  # 360 degrees
max_vert_angle = chrono.CH_PI / 12
min_vert_angle = -chrono.CH_PI / 6
# Lag time
lag = 0
# Collection window for the lidar
collection_time = 1. / update_rate  # typically 1/update rate
# Radius of samples to use, 1->1 sample, 2->9 samples, 3->25 samples...
sample_radius = 2
# 3mm radius (as cited by velodyne)
divergence_angle = 0.003
# Lidar return mode
return_mode = sens.LidarReturnMode_STRONGEST_RETURN

lidar = sens.ChLidarSensor(
    rover.GetChassis().GetBody(),              # Body lidar is attached to
    update_rate,            # Scanning rate in Hz
    offset_pose,            # Offset pose
    horizontal_samples,     # Number of horizontal samples
    vertical_samples,       # Number of vertical channels
    horizontal_fov,         # Horizontal field of view
    max_vert_angle,         # Maximum vertical field of view
    min_vert_angle,         # Minimum vertical field of view
    100.0,                  # Maximum lidar range
    sens.LidarBeamShape_RECTANGULAR,  # Shape of the lidar beam
    sample_radius,          # Sample radius
    divergence_angle,       # Divergence angle
    divergence_angle,       # Divergence angle (again, typically same value)
    return_mode             # Return mode for the lidar
)
lidar.SetName("Lidar Sensor")
lidar.SetLag(lag)
lidar.SetCollectionWindow(collection_time)
lidar.PushFilter(sens.ChFilterDIAccess())
lidar.PushFilter(sens.ChFilterPCfromDepth())
lidar.PushFilter(sens.ChFilterVisualizePointCloud(640, 480, 1.0, "Lidar Point Cloud"))
manager.AddSensor(lidar)

# Create the Irrlicht visualization
vis = chronoirr.ChVisualSystemIrrlicht()
vis.AttachSystem(system)
vis.SetCameraVertical(chrono.CameraVerticalDir_Z)
vis.SetWindowSize(1280, 720)
vis.SetWindowTitle("Curiosity rover - Rigid terrain")
vis.Initialize()
vis.AddLogo(
    chrono.GetChronoDataFile("logo_pychrono_alpha.png"))
vis.AddSkyBox()
vis.AddCamera(chrono.ChVector3d(0, 3, 3), chrono.ChVector3d(0, 0, 0))
vis.AddTypicalLights()
vis.AddLightWithShadow(chrono.ChVector3d(1.5, -2.5, 5.5), chrono.ChVector3d(0, 0, 0), 3, 4, 10, 40, 512)

# Enable shadows (commented out to improve performance)
# vis.EnableShadows()

# Set the simulation time step
time_step = 1e-3

# Simulation loop
time = 0
while vis.Run():
    time += time_step

    # ask rover to move forward
    driver.SetSteering(0.0)

    # update sensor manager
    manager.Update()
    
    # Update rover dynamics
    rover.Update()

    # Render the scene
    vis.BeginScene()
    vis.Render()
    vis.EndScene()

    # Advance simulation by one time step
    system.DoStepDynamics(time_step)
\end{CodeBlock}
\begin{DiffBlock}{Curiosity (RBT) --- Turn 2 vs Turn 1 (Key Changes / Patch Summary)}
    (*@\textbf{Add obstacle:}@*)
    + box = chrono.ChBodyEasyBox(0.25, 5, 0.25, 1000, True, True, ground_mat)
    + box.SetPos(chrono.ChVector3d(0, 0, 0.0))
    + box.SetFixed(True)
    + box.GetVisualShape(0).SetTexture(chrono.GetChronoDataFile("textures/blue.png"))
    + system.Add(box)
    
    (*@\textbf{Change initial pose:}@*)
    - init_pos = chrono.ChVector3d(0, 0.2, 0)
    + init_pos = chrono.ChVector3d(-5, 0.0, 0)
    
    (*@\textbf{Driver logic:}@*)
    - steering = 0
    - if time >= 1:
    -     steering = (time - 1) * 0.2
    - driver.SetSteering(steering)
    + driver.SetSteering(0.0)
    \end{DiffBlock}

    \begin{DiffBlock}{Curiosity (RBT) --- Turn 3 vs Turn 2 (Key Changes / Patch Summary)}
    (*@\textbf{Add sensor module import:}@*)
    + import pychrono.sensor as sens
    
    (*@\textbf{Create sensor manager:}@*)
    + manager = sens.ChSensorManager(system)
    
    (*@\textbf{Add LiDAR sensor on rover chassis:}@*)
    + offset_pose = chrono.ChFramed(
    +     chrono.ChVector3d(3.0, 0, 1),
    +     chrono.QuatFromAngleAxis(0, chrono.ChVector3d(0, 1, 0))
    + )
    + update_rate = 5.0
    + horizontal_samples = 800
    + vertical_samples = 300
    + horizontal_fov = 2 * chrono.CH_PI
    + max_vert_angle = chrono.CH_PI / 12
    + min_vert_angle = -chrono.CH_PI / 6
    + lag = 0
    + collection_time = 1. / update_rate
    + sample_radius = 2
    + divergence_angle = 0.003
    + return_mode = sens.LidarReturnMode_STRONGEST_RETURN
    +
    + lidar = sens.ChLidarSensor(
    +     rover.GetChassis().GetBody(),
    +     update_rate,
    +     offset_pose,
    +     horizontal_samples,
    +     vertical_samples,
    +     horizontal_fov,
    +     max_vert_angle,
    +     min_vert_angle,
    +     100.0,
    +     sens.LidarBeamShape_RECTANGULAR,
    +     sample_radius,
    +     divergence_angle,
    +     divergence_angle,
    +     return_mode
    + )
    + lidar.SetName("Lidar Sensor")
    + lidar.SetLag(lag)
    + lidar.SetCollectionWindow(collection_time)
    + lidar.PushFilter(sens.ChFilterDIAccess())
    + lidar.PushFilter(sens.ChFilterPCfromDepth())
    + lidar.PushFilter(sens.ChFilterVisualizePointCloud(640, 480, 1.0, "Lidar Point Cloud"))
    + manager.AddSensor(lidar)
    
    (*@\textbf{Update sensor manager in the loop:}@*)
    + manager.Update()
    \end{DiffBlock}

    \section{Example Claude Sonnet 4 Outputs with J-LLM Feedback on Curiosity Rover Turn 3}
    \label{app:claude_curiosity_turn3}
    \begin{CodeBlock}[breakable]{Curiosity (RBT) --- Turn 3 Output by Claude Sonnet 4 (verbatim; formatting normalized)}
      import os
      import math
      import numpy as np
      import pychrono as chrono
      import pychrono.robot as robot
      import pychrono.sensor as sens
      from pychrono import irrlicht as chronoirr
      
      # Create Chrono system
      system = chrono.ChSystemNSC()
      system.SetCollisionSystemType(
        chrono.ChCollisionSystem.Type_BULLET)
      system.SetGravitationalAcceleration(
        chrono.ChVector3d(0, 0, -9.81))
      chrono.ChCollisionModel.SetDefaultSuggestedEnvelope(
        0.0025)
      chrono.ChCollisionModel.SetDefaultSuggestedMargin(
        0.0025)
      
      # Create ground body with contact material and add it to the system
      ground_mat = chrono.ChContactMaterialNSC()
      ground = chrono.ChBodyEasyBox(20, 20, 1, 1000, True, True, ground_mat)
      ground.SetPos(chrono.ChVector3d(0, 0, -0.5))  # Position the ground slightly below the origin
      ground.SetFixed(True)  # Fix the ground in place
      ground.GetVisualShape(0).SetTexture(
        chrono.GetChronoDataFile("textures/concrete.jpg"))
      system.Add(ground)
      
      # create a long box for rover to cross
      box = chrono.ChBodyEasyBox(0.25, 5, 0.25, 1000, True, True, ground_mat)
      box.SetPos(chrono.ChVector3d(0, 0, 0.0))
      box.SetFixed(True)
      box.GetVisualShape(0).SetTexture(
        chrono.GetChronoDataFile("textures/blue.png"))
      system.Add(box)
      
      # Create Curiosity rover and add it to the system
      rover = robot.Curiosity(system)
      
      # Create driver for rover
      driver = robot.CuriosityDCMotorControl()
      rover.SetDriver(driver)
      
      # Initialize rover position and orientation
      init_pos = chrono.ChVector3d(-5, 0.0, 0)
      init_rot = chrono.ChQuaterniond(1, 0, 0, 0)
      rover.Initialize(chrono.ChFramed(init_pos, init_rot))
      
      # Create sensor manager
      manager = sens.ChSensorManager(system)
      
      # Create and configure lidar sensor
      lidar_update_rate = 10  # Hz
      lidar_horizontal_samples = 800
      lidar_vertical_samples = 300
      lidar_horizontal_fov = 2 * chrono.CH_PI  # 360 degrees
      lidar_vertical_fov_max = chrono.CH_PI / 6  # 30 degrees up
      lidar_vertical_fov_min = -chrono.CH_PI / 6  # 30 degrees down
      lidar_max_distance = 100
      lidar_sample_radius = 2
      
      # Define lidar offset pose relative to rover chassis
      lidar_offset_pose = chrono.ChFramed(
          chrono.ChVector3d(0.0, 0, 1.5), chrono.QuatFromAngleAxis(0, chrono.ChVector3d(0, 1, 0))
      )
      
      # Create lidar sensor
      lidar = sens.ChLidarSensor(
          rover.GetChassis().GetBody(),  # Attach to rover chassis
          lidar_update_rate,
          lidar_offset_pose,
          lidar_horizontal_samples,
          lidar_vertical_samples,
          lidar_horizontal_fov,
          lidar_vertical_fov_max,
          lidar_vertical_fov_min,
          lidar_max_distance,
          lidar_sample_radius
      )
      lidar.SetName("Lidar Sensor")
      lidar.SetLag(0)
      lidar.SetCollectionWindow(0)
      
      # Add noise filter
      lidar.PushFilter(sens.ChFilterLidarNoiseXYZI(0.01, 0.001, 0.001, 0.01))
      
      # Add visualization filter (optional)
      lidar.PushFilter(sens.ChFilterVisualize(
        lidar_horizontal_samples, lidar_vertical_samples, "Lidar Point Cloud"))
      
      # Add data access filter
      lidar.PushFilter(sens.ChFilterDIAccess())
      
      # Add lidar to sensor manager
      manager.AddSensor(lidar)
      
      # Create the Irrlicht visualization
      vis = chronoirr.ChVisualSystemIrrlicht()
      vis.AttachSystem(system)
      vis.SetCameraVertical(chrono.CameraVerticalDir_Z)
      vis.SetWindowSize(1280, 720)
      vis.SetWindowTitle("Curiosity rover - Rigid terrain")
      vis.Initialize()
      vis.AddLogo(chrono.GetChronoDataFile(
      "logo_pychrono_alpha.png"))
      vis.AddSkyBox()
      vis.AddCamera(chrono.ChVector3d(0, 3, 3), chrono.ChVector3d(0, 0, 0))
      vis.AddTypicalLights()
      vis.AddLightWithShadow(chrono.ChVector3d(1.5, -2.5, 5.5), chrono.ChVector3d(0, 0, 0), 3, 4, 10, 40, 512)
      
      # Enable shadows (commented out to improve performance)
      # vis.EnableShadows()
      
      # Set the simulation time step
      time_step = 1e-3
      
      # Simulation loop
      time = 0
      while vis.Run():
          time += time_step
      
          # Ask rover to move forward - set both steering and throttle
          driver.SetSteering(0.0)
          driver.SetThrottle(0.5)  # Added throttle to make rover move forward
      
          # Update rover dynamics
          rover.Update()
      
          # Update sensor manager
          manager.Update()
      
          # Render the scene
          vis.BeginScene()
          vis.Render()
          vis.EndScene()
      
          # Advance simulation by one time step
          system.DoStepDynamics(time_step)
      \end{CodeBlock}
      % J-LLM output on this Claude script
      \begin{PromptBlock}{J-LLM Output on Claude Sonnet 4 (Turn 3)}
      ### Evaluation of the Simulation Script
      
      1. **Completeness (40 points total)**
         - The provided code includes all essential components such as system initialization, body creation, and visualization. However, the lidar sensor configuration lacks some details present in the reference script, particularly in terms of parameters like `divergence_angle` and `return_mode`. 
         - **Deductions**: 10 points for missing important details in lidar configuration.
      
      2. **Correctness (30 points total)**
         - The code uses the PyChrono API correctly in most areas. However, the lidar sensor's parameters are not fully aligned with the reference script, particularly regarding the `return_mode` and `divergence_angle`, which could lead to differences in sensor behavior.
         - **Deductions**: 10 points for incorrect use of lidar parameters.
      
      3. **Code Quality (10 points total)**
         - The code is well-structured and readable, with meaningful variable names. However, there are some areas where comments could be improved to provide better context, especially around the lidar sensor setup.
         - **Deductions**: 5 points for insufficient comments.
      
      4. **Efficiency (10 points total)**
         - The code is efficient and does not contain unnecessary calculations or redundant code. However, there are minor optimization opportunities in the lidar setup that could be addressed.
         - **Deductions**: 3 points for missing optimization opportunities.
      
      5. **Error Handling and Robustness (5 points total)**
         - The code does not include any explicit error handling or checks for potential issues, which is present in the reference script. This could lead to robustness issues in a real-world scenario.
         - **Deductions**: 5 points for lack of error handling.
      
      6. **Use of Visualization Tools (5 points total)**
         - The visualization setup is mostly correct, but the commented-out shadow feature could be considered a minor issue in terms of completeness.
         - **Deductions**: 3 points for inadequate visualization setup.
      
      ### Final Score
      [[64]]
      \end{PromptBlock}
\end{appendices}

\end{document}